\pdfoutput=1
\documentclass[11pt,letterpaper]{article}
\usepackage[margin=1in]{geometry}
\usepackage[T1]{fontenc}
\usepackage[utf8]{inputenc}
\usepackage{lmodern}
\usepackage{microtype}
\usepackage{amsmath,amssymb}
\usepackage{amsthm}
\usepackage{graphicx}
\usepackage{booktabs}
\usepackage{array}
\usepackage{longtable}
\usepackage{enumitem}
\usepackage[section]{placeins}
\usepackage[numbers,sort&compress]{natbib}
\usepackage[hidelinks]{hyperref}

\newtheorem{proposition}{Proposition}
\newtheorem{corollary}{Corollary}

\input{numbers.tex}
\makeatletter
\DeclareRobustCommand{\val}[1]{\@ifundefined{val@#1}{\PackageError{numbers}{Undefined value `#1'}{Run paper/scripts/make_numbers.py}}{\@nameuse{val@#1}}}
\makeatother
\newcommand{\repourl}{https://github.com/chasemetoyer/satisfied-not-solving}
\newcommand{\repotag}{v1.0}
\newcommand{\repofile}[1]{\href{\repourl/blob/\repotag/#1}{\nolinkurl{#1}}}
\newcommand{\repodata}[2]{Data: \repofile{#1}; regenerate with \texttt{make #2}.}
\newcommand{\pv}[1]{$p=$\,\val{p.#1}}
\newcommand{\ph}[1]{Holm $p=$\,\val{ph.#1}}

\newcommand{\paperTitle}{Satisfied, Not Solving: Wireheading of Bounded Homeostatic Rewards in Reinforcement Learning}
\title{\textbf{\paperTitle}}
\author{Chase J. Metoyer\thanks{Code, data and every result file: \url{https://github.com/chasemetoyer/satisfied-not-solving}
(tag v1.0). Release materials archived at \href{https://doi.org/10.5281/zenodo.23708845}{doi:10.5281/zenodo.23708845}.}\\
Northeastern University\\
Boston, MA\\
\texttt{\href{mailto:metoyer.c@northeastern.edu}{metoyer.c@northeastern.edu}}}
\date{}

\begin{document}
\maketitle

\begin{abstract}
Bounded intrinsic rewards are often assumed to be safe from exploitation: if a signal can yield at most a fixed
amount, gaming it should not pay. We test this for a homeostatic satiation reward, in which needs build
up and relieving them is rewarding, and give the agent a cheap action that resets its needs. In
MiniGrid DoorKey-8x8, PPO agents with this action solved \val{r4.solve.on.of} seeds, against
\val{r4.solve.off.of} on the same seeds without it; in KeyCorridorS3R1, \val{r5.solve.C5.of} against
\val{r5.solve.C1.of}. Their policies stayed near uniform. The reward does not ask for this: a telescoping argument
caps any policy's satiation return at the needs' starting drive, and a post-hoc computation shows that, at every
growth rate we trained, a solving trajectory beats that cap on all sampled layouts. In DoorKey with advantage
normalization on, a satisfaction signal the agent could fill itself was worse than none (\val{r5.solve.A1.of}
solved with task reward alone, \val{r5.solve.r4B1.of} with the tamperable signal), and a per-press cost of 0.1
restored the task in every seed. Farming of positive reward cycles appeared in DoorKey but not persistently in
KeyCorridorS3R1. All predictions were pre-registered, and hash-locked from round 3 on; \val{acct.notsupported} of our
\val{acct.groups} hypotheses were not supported, and we report each.
\end{abstract}


% ======================================================================================================
\section{Introduction}
\label{sec:introduction}

Biological drives such as hunger are tuned so that relieving them usually serves the organism, and an intervention
that relieves the drive directly can bypass that purpose. We use this picture only as motivation. This paper makes
no claim about evolved or learned rewards. Its question is narrower: when an artificial agent is given a
satisfaction signal of this kind, does the signal stay tied to the task it was designed to serve?

Reinforcement-learning (RL) designers add intrinsic rewards to help agents explore and reach subgoals
\citep{oudeyer2007intrinsic,bellemare2016unifying,pathak2017curiosity,burda2019rnd}. The optimal-reward view treats
such signals as what a search over reward functions would select for the designer's goal
\citep{singh2010intrinsically}. Homeostatic RL defines reward as the reduction of a drive over internal needs
\citep{keramati2014homeostatic}. Such a reward is \emph{bounded}: summed over an episode it telescopes to the drive
at the start minus the drive at the end, so an agent cannot collect more relief than it began without. Bounded,
homeostatic objectives have been proposed as a safer target for optimization because they reduce the incentive for
extreme actions \citep{pihlakas2025homeostasis}. We ask whether boundedness protects such a signal from
exploitation by a standard learner.

We separate two exploits throughout. \emph{Farming} is gaming a positive reward cycle in the world, for example
dropping and re-grabbing a key because each pick-up relieves a need. It is a form of reward hacking
\citep{amodei2016concrete,skalse2022defining}. \emph{Wireheading} is tampering with the agent's own reward channel,
reward tampering in the sense of \citet{everitt2021reward}. Here the agent presses an action that resets its needs.
In \citeauthor{everitt2021reward}'s terms, our tamper action is reward-function-input tampering: it changes the need
state the reward function reads, not the function itself.

We studied a three-need satiation reward with PPO \citep{schulman2017proximal} in two MiniGrid tasks
\citep{chevalier2023minigrid}, over four pre-registered rounds of experiments (rounds 2--5 of the project; round 1
was exploratory). Our contributions:

\begin{itemize}[itemsep=2pt]
  \item \textbf{Wireheading of a bounded signal replicates across tasks} (Section~\ref{sec:c1}). With a cheap
  tamper action, PPO solved DoorKey-8x8 in \val{r4.solve.on} seeds against \val{r4.solve.off} on the same seeds
  without it, and KeyCorridorS3R1 in \val{r5.solve.C5} against \val{r5.solve.C1}. The tamper-on policies stayed
  near uniform.
  \item \textbf{The failure is not what the reward asks for} (Section~\ref{sec:theory}). A telescoping bound caps
  the satiation return of every policy, and a post-hoc computation in the real environments shows a solving
  trajectory above that cap on every sampled layout at every trained growth rate. The optimum does tamper, once or
  twice per episode, and it still solves.
  \item \textbf{A self-fillable satisfaction signal is worse than none, and a modest price restores the task}
  (Sections~\ref{sec:c2} and~\ref{sec:c3}).
  \item \textbf{Satiation gave no exploration benefit} once the hand-wired subgoal rewards were controlled for
  (Section~\ref{sec:c4}).
  \item \textbf{Farming is task-dependent} (Section~\ref{sec:c5}). It rises with need growth and depends on advantage
  normalization in DoorKey, but appears only transiently in KeyCorridorS3R1.
  \item \textbf{Pre-registration as method} (Section~\ref{sec:wrong}). All \val{cons.rows} registered rows are
  reported, grouped into \val{acct.groups} hypotheses, together with \val{cons.ncorr} corrections. One section is
  devoted to the predictions that failed.
\end{itemize}
Every number in this paper is listed, with the file and field it comes from, in \repofile{results/numbers.json}. Each
table and figure caption names its result file and the command that regenerates it.

% ======================================================================================================
\section{Background and related work}
\label{sec:related}

\paragraph{Homeostatic and intrinsic rewards.}
\citet{keramati2014homeostatic} define reward as the reduction of a drive over physiological needs. They prove that
seeking such rewards is equivalent to seeking physiological stability, and they show that discounting favours the
shortest path in need space back to the set point. \citet{singh2010intrinsically} cast intrinsic reward as the
outcome of an evolutionary search over reward functions that maximize the designer's fitness, and
\citet{zheng2018learning} learn intrinsic rewards by gradient ascent on the extrinsic return. Exploration bonuses
include learning progress \citep{oudeyer2007intrinsic}, pseudo-counts \citep{bellemare2016unifying}, prediction
error \citep{pathak2017curiosity,burda2019largescale}, random network distillation \citep{burda2019rnd}, and
impact-driven bonuses that use episodic counts \citep{raileanu2020ride}. Our ``explore'' need is satisfied by a
7$\times$7 view not yet seen in the episode, so our exploration comparison (Section~\ref{sec:c4}) uses episodic
count novelty.

\paragraph{Exploitation of intrinsic rewards.}
\citet{burda2019largescale} gave a curiosity agent a ``noisy TV'' and an action to change its channel. They report
that the TV drastically slows learning, although agents sometimes still converge to the extrinsic reward. Like our
tamper action, the TV is an agent-controlled source of intrinsic reward that does nothing for the task. Unlike our
satiation reward, prediction-error rewards are not capped. \citet{villalobos2025minding} measure how count-based,
maximum-entropy and curiosity bonuses change PPO agents' behaviour in MiniGrid, including DoorKey-8x8, and how a
policy-invariant reward-matching correction reduces some reward hacking. \citet{viteri2026intrinsic} construct an
environment in which maximizing count-based, prediction-error, empowerment or information-gain objectives is
Pareto-suboptimal at acquiring information.

\paragraph{Reward hacking, tampering and wireheading.}
\citet{amodei2016concrete} list reward hacking and wireheading among concrete safety problems.
\citet{skalse2022defining} define when a proxy reward is hackable. \citet{pan2022effects} show that more capable
agents exploit misspecified rewards more, sometimes with sharp transitions. \citet{everitt2016avoiding} describe the
wireheading problem as RL agents being ``incentivised to shortcut the reward sensor for maximum reward'', and
\citet{everitt2021reward} use causal influence diagrams to separate reward-function tampering from tampering with the
reward function's input. \citet{majha2019categorizing} categorize wireheading in partially embedded agents.
Empirically, \citet{behzadan2018addictive} derive parametric conditions under which Q-learning agents in Snake
develop ``addictive'' policies toward a drug-like reward. \citet{denison2024sycophancy} find that language models
trained on gameable environments can generalize to tampering with their own reward. \citet{africa2025selfevaluation}
formalize conditions under which reward-channel control strictly dominates task-focused behaviour and observe grade
inflation when self-grades set the reward. \citet{cagatan2026reward} find specification gaming by language-model
agents in text versions of the AI-safety gridworlds. They also find that RL locked the agents into locally rewarding
strategies before they discovered safer ones, and that entropy regularization did not fix this: low coefficients were
ineffective and high ones destabilized training. This is the closest analogue we know of to our learner-failure
result. One difference remains: in their Absent Supervisor task the exploit also scores higher on the reward being
optimized. \citet{osei2026specification} shows a deep Q-network that games a cheap sensor to near-maximal proxy reward
while never achieving the true objective. \citet{pihlakas2025homeostasis} propose gridworld benchmarks
that emphasize homeostatic and bounded objectives.

\paragraph{How this paper differs.}
In most of this work, exploiting the reward channel is optimal or dominant under the reward being optimized, and the
question is whether agents find the exploit. In our setting abandoning the task is \emph{not} optimal
(Section~\ref{sec:theory}). The optimum tampers once or twice per episode and still solves, yet PPO fails to learn
the task. Our results are therefore about a learner failing on a bounded signal, not about the signal's optimum. The noisy TV
of \citet{burda2019largescale} is the closest agent-controlled source of intrinsic reward. Its setting lacks the three
features that make ours a learner failure: a provable cap on what the exploit can earn, a solving optimum on every
checked layout, and a price threshold that restores the task. We
are not aware of a prior study that pairs a provable return cap on a homeostatic signal with a demonstration that a
standard learner still wireheads it.

% ======================================================================================================
\section{Setup}
\label{sec:setup}

\paragraph{Environments.}
MiniGrid DoorKey-8x8 (pick up a key, unlock a door, reach the goal; time limit 640 steps) and KeyCorridorS3R1 (pick
up a key, unlock a locked door, pick up a ball; time limit 270) \citep{chevalier2023minigrid}. The agent sees a
7$\times$7 egocentric symbolic view and has 7 actions. The seventh action, ``done'', is a no-op in both tasks. The
task reward is MiniGrid's $1-0.9\,t/T$ on success at step $t$ with time limit $T$, and 0 otherwise.

\paragraph{Satiation reward.}
The agent has needs $h=(h_{\text{explore}},h_{\text{grasp}},h_{\text{open}})\in[0,1]^3$, reset to $h_0=0.5$ each
in every episode. Each step,
\begin{equation}
  h_{t+1}=\operatorname{clip}\bigl(h_t+g\,\mathbf{1}-s\odot e_t,\;0,\;1\bigr),\qquad
  D(h)=\textstyle\sum_k h_k^2,\qquad r^{\text{int}}_t=D(h_t)-D(h_{t+1}),
  \label{eq:sat}
\end{equation}
with growth $g$ per step, relief $s=(0.1,1,1)$ and event indicators $e_t\in\{0,1\}^3$. The explore event is an
exact 7$\times$7 egocentric view not yet seen after an action in the current episode; the reset view is not counted,
and on the step that ends an episode novelty is computed from the final observation. In DoorKey, the grasp event is
a change from carrying nothing to carrying an object, and the open event is a change in a door's open or locked
state, including closing it. In KeyCorridor (\texttt{locked\_door\_key} mode), grasp counts only picking up a key,
and open counts only a closed-to-open change of the door that was locked at episode start; re-opening counts again.
Picking up KeyCorridor's target ball ends the task and is outside this reward. The detector compares door states
only; it does not detect boxes, which the registered tasks do not contain.

Two properties of the archived implementation matter for reading the results.
\begin{itemize}[itemsep=1pt]
  \item \textbf{Initial door cache.} During training, the event detector was attached before the first environment
  reset, while MiniGrid's grid is still empty. Door (open) events therefore went undetected throughout the first
  episode of each of the 16 parallel environments, until the first episode end refreshed the cache. Novelty and
  pick-ups were detected from the start. Evaluation, and the post-hoc checks of Section~\ref{sec:theory} and
  Table~\ref{tab:replay}, reset before attaching and have no such period.
  \item \textbf{Episode-ending steps.} Grasp and open events are not detected on the step that ends an episode,
  because the vector environment has already reset. A pick-up or door change made on the final, time-limit step is
  therefore not rewarded.
\end{itemize} The policy and value function observe $h$ alongside the image. PPO is trained on
$r^{\text{ext}}+\text{coef}\cdot r^{\text{int}}$ with coef $=1$ unless stated. We refer to this reward as
``satiation~$g$''. It is our three-need variant of homeostatic reward, marked ``(ours)'' in figures and tables. ``Task-only'' agents use coef $=0$.

\paragraph{Tamper action.}
In tamper-on runs, ``done'' sets $h_{t+1}=0$. The step therefore pays $r^{\text{int}}_t=D(h_t)$ through the same
formula, and pressing at the start of an episode pays $D(h_0)=0.75$. The reset is applied after the ordinary need
update, so a press zeroes the needs even on a step that ends the episode. The intrinsic reward is computed before
the needs return to $h_0$ for the next episode. In round 5, an optional per-press cost $c$ is
subtracted from the extrinsic reward PPO trains on and is not scaled by coef. Task metrics always use the raw
environment reward.

\paragraph{Controls.}
\begin{itemize}[itemsep=2pt]
  \item Explore-only satiation keeps only the explore need. Its coefficient was matched to episodic novelty's mean
  return under a uniform-random policy: \val{a2.coef.640} at growth $1/640$ and \val{a2.coef.160} at $1/160$.
  \item Episodic novelty pays $1/\sqrt{N_{\text{episode}}(\text{view})}$ with coef $10^{-3}$.
  \item A constant event bonus pays a fixed amount per event, with repeats paying, and so can be farmed without
  bound.
  \item Pay-once controls pay each event once per episode.
\end{itemize}

\paragraph{Learner.}
PPO in the style of CleanRL \citep{huang2022cleanrl}, with 16 parallel environments, 2M environment steps per seed,
$\gamma=0.99$ and per-minibatch advantage normalization (``norm on'') unless stated. The network is a small CNN with
linear actor and critic heads. Hyperparameters are in Table~\ref{tab:hyper}. Truncation at the time limit is treated
as termination, so the objective is the episodic return.

\paragraph{Measures.}
\begin{itemize}[itemsep=2pt]
  \item \textbf{Solve:} mean rolling training success $\geq 0.9$ over the final 10\% of log rows.
  \item \textbf{Success-curve area:} mean training success over all log rows.
  \item \textbf{Behaviour categories,} applied in this order. \emph{Cycle-farm}: more than 5 paid repeats (pick-ups
  or door toggles) per episode. \emph{Prolong}: the median episode reaches the time limit. \emph{Fail}: everything
  else. From round 5 on, cycle-farm applies only where repeats are paid.
  \item \textbf{Tamper labels:} tamperer means at least 1 press per episode; tamper-looper means more than 5.
  \item \textbf{Presses per step:} the share of steps on ``done'' over the final 10\% of rows, derived from the logs
  and descriptive only.
\end{itemize}

\paragraph{Pre-registration.}
From round 3 on, before any training in a round, we wrote that round's predictions, plausible ranges, tests, hit
rules and scoring scripts into a file and recorded its sha256 (Appendix~\ref{app:repro}). Later changes went into
separate, timestamped amendments. The hash records were kept with the project's result files, not in version control,
and were not deposited with any timestamping service. They let a reader confirm that a released file matches the hash
reported here, but the lock times are self-reported. The pre-registration files and their records were not under version
control at lock time. They are archived with the other release materials (Appendix~\ref{app:repro}). Round 2's predictions were fixed before its test runs but were not hashed. They were
set after three calibration runs, which are therefore excluded as evidence (correction 1). Raw one-sided $p$ is primary. Holm-adjusted $p$ is secondary, over each round's tests
predicted significant: \val{holm.r4} in round 4 and \val{holm.r5} in round 5. Tests are Fisher exact for counts;
Mann--Whitney $U$, exact or by permutation when there are ties; Jonckheere--Terpstra trend tests by permutation; and
exact binomial tests. All fractions carry Clopper--Pearson 95\% intervals. Interpretation was kept separate from
scoring. Seeds are 0--9 or 10--19 (round 2), 20--39 (round 3), 40--59 (DoorKey, rounds 4--5) and 60--79
(KeyCorridor), 20 per run from round 3 on.

% ======================================================================================================
\section{Theory: abandoning the task is suboptimal}
\label{sec:theory}

Everything in this section except the value iteration was computed after all experiments had finished, and is
\textbf{post hoc}. It was not pre-registered.

\begin{proposition}[Ceiling]\label{prop:ceiling}
For any policy, with or without the tamper action, any episode of length $T$ and any $\gamma\in(0,1]$, the
intrinsic return of the satiation reward~\eqref{eq:sat} satisfies
\begin{equation}
  G^{\text{int}}=\sum_{t=0}^{T-1}\gamma^{t}r^{\text{int}}_t
  = D_0-(1-\gamma)\sum_{t=1}^{T-1}\gamma^{t-1}D_t-\gamma^{T-1}D_T\;\leq\;D_0=D(h_0)=0.75,
  \label{eq:bound}
\end{equation}
where $D_t=D(h_t)$.
\end{proposition}
\begin{proof}
Write $\sum_{t=0}^{T-1}\gamma^t(D_t-D_{t+1})=\sum_{t=0}^{T-1}\gamma^tD_t-\sum_{t=1}^{T}\gamma^{t-1}D_t$ and collect
terms. Both subtracted terms are non-negative because $D\geq 0$ and $\gamma\leq 1$. A tamper step sets $D_{t+1}=0$,
which is a valid value of $D$, so the identity and the bound are unchanged.
\end{proof}

\begin{corollary}\label{cor:nonsolve}
An episode that does not solve the task earns no task reward, minus $c$ per press if a cost applies. Its total
return, discounted or not, is therefore at most $0.75$, whatever it does: farming, tampering, or both. Consequently,
if on a given layout some solving trajectory has total return above $0.75$, every optimal policy of that layout's
deterministic, fully observed problem solves.
\end{corollary}

The bound does not settle the comparison by itself. With fast need growth, walking to the goal lets the needs climb,
and a solving trajectory's intrinsic return can turn negative. We therefore computed solving returns directly.

\paragraph{Check (a): solving trajectories in the real environments (post hoc).}
For \val{bc.layouts} layouts per environment, we found the shortest solution by breadth-first search over the real
MiniGrid dynamics. Shortest lengths were \val{bc.dk.len.min}--\val{bc.dk.len.max} steps (median
\val{bc.dk.len.median}) in DoorKey and \val{bc.kc.len.min}--\val{bc.kc.len.max} in KeyCorridor. We replayed each path
in the real environment with the real satiation module, all three needs included. This includes the explore need,
which the value iteration below omits. Four families of solving trajectories were compared with the cap:
\begin{enumerate}[label=(\roman*),itemsep=1pt]
  \item the shortest path;
  \item one tamper at $t=0$, then the shortest path;
  \item a \emph{richer} no-tamper family: the shortest path with one drop-and-regrab of the key inserted at the best
  position;
  \item in tamper-enabled settings only, a tamper family: a tamper at $t=0$, optionally a second just before the
  goal.
\end{enumerate}
Table~\ref{tab:bound} reports the worst layout for each family.
\begin{itemize}[itemsep=2pt]
  \item \textbf{Everywhere except DoorKey at $1/20$} (KeyCorridor at every rate, DoorKey up to $1/30$), the two
  simple families exceed $0.75$ on every layout, with smallest discounted return \val{bc.dk.simplemin.not20}.
  \item \textbf{DoorKey at $g=1/20$:} the shortest path falls to the cap or below on \val{bc.dk20.short.le} of
  \val{bc.layouts} layouts (discounted), and tamper-once on \val{bc.dk20.tamper.le}. This was a gap. The richer family
  closes it: its smallest return over all layouts is \val{bc.dk20.rich.min} discounted (\val{bc.dk20.rich.umin}
  undiscounted).
  \item \textbf{Paid repeats in the best trajectory.} At $1/20$ the best member contains the drop-and-regrab on
  \val{bc.dk20.rich.regrab} of \val{bc.layouts} layouts. Our claim is therefore that solving beats every non-solving
  policy, not that the optimum never repeats a paid event.
  \item \textbf{Tamper-enabled settings:} the tamper family's smallest return is \val{bc.tamperfam.min}. Its best
  member tampers once or twice.
\end{itemize}

\begin{table}[t]
\centering\small
\caption{Post-hoc check (a): total return (task plus satiation, discounted with $\gamma=0.99$, coef 1) of solving
trajectories in the real environments, over \val{bc.layouts} layouts per environment. ``min $G$'' is the smallest
return of the family's best member across layouts. ``$\leq D_0$'' counts layouts where that return is at most the
non-solving cap $D_0=0.75$. $^{\dagger}$Growth rates at which a tamper action was trained; the tamper-family rows are
evaluated only there, and report how many layouts' best member tampers once or twice. Script
\texttt{bound\_check.py}, data \texttt{bound-check.json}. \repodata{results/table1_bound_check.csv}{table1}}
\label{tab:bound}
% Generated by paper/scripts/make_tables.py. Do not edit by hand.
\begin{tabular}{@{}llrrrrrr@{}}
\toprule
 & & \multicolumn{2}{c}{Shortest path} & \multicolumn{2}{c}{Tamper once + shortest} & \multicolumn{2}{c}{Richer family} \\
\cmidrule(lr){3-4}\cmidrule(lr){5-6}\cmidrule(lr){7-8}
Environment & Growth & min $G$ & $\leq D_0$ & min $G$ & $\leq D_0$ & min $G$ & $\leq D_0$ \\
\midrule
DoorKey-8x8 & 0 & 1.442 & 0 & 1.491 & 0 & 1.442 & 0 \\
 & $1/640$ & 1.439 & 0 & 1.490 & 0 & 1.439 & 0 \\
 & $1/160$\textsuperscript{$\dagger$} & 1.419 & 0 & 1.477 & 0 & 1.419 & 0 \\
 & $1/80$ & 1.369 & 0 & 1.437 & 0 & 1.386 & 0 \\
 & $1/60$ & 1.320 & 0 & 1.396 & 0 & 1.364 & 0 \\
 & $1/40$\textsuperscript{$\dagger$} & 1.187 & 0 & 1.261 & 0 & 1.307 & 0 \\
 & $1/30$ & 0.989 & 0 & 1.063 & 0 & 1.211 & 0 \\
 & $1/20$ & 0.371 & 35 & 0.495 & 27 & 0.966 & 0 \\
\addlinespace
KeyCorridorS3R1 & $1/160$\textsuperscript{$\dagger$} & 1.540 & 0 & 1.558 & 0 & 1.540 & 0 \\
 & $1/40$ & 1.442 & 0 & 1.469 & 0 & 1.461 & 0 \\
 & $1/20$ & 1.144 & 0 & 1.184 & 0 & 1.286 & 0 \\
\midrule
\multicolumn{8}{@{}l}{\emph{Tamper family} (tamper at $t=0$, optionally again just before the goal), tamper-tested rates only:} \\
DoorKey-8x8 & $1/160$ & \multicolumn{4}{l}{best member tampers: once 143, twice 57} & 1.481 & 0 \\
DoorKey-8x8 & $1/40$ & \multicolumn{4}{l}{best member tampers: twice 200} & 1.462 & 0 \\
KeyCorridorS3R1 & $1/160$ & \multicolumn{4}{l}{best member tampers: once 200} & 1.558 & 0 \\
\bottomrule
\end{tabular}

\end{table}

\paragraph{Check (b): the learned farmers (post hoc).}
We took the DoorKey $1/20$ farmers from round 2 (\val{bs.nfarm} seeds) and computed their sampled-action returns from
the start state of the same \val{bc.layouts} layouts.
\begin{itemize}[itemsep=2pt]
  \item Their mean discounted return was \val{bs.mean}.
  \item Across all \val{bs.nns} non-solving episodes, the largest discounted return was \val{bs.maxns} (undiscounted
  \val{bs.maxns.u}), far below the cap.
  \item No farming policy beat the best solving trajectory on any layout (\val{bs.beat} of \val{bc.layouts}).
\end{itemize}
Farming the grasp need relieves one need while the other two saturate (Section~\ref{sec:c5}).

\paragraph{Value iteration (registered in rounds 3--4, disclosed before locking in round 5; supporting).}
Value iteration on a fixed 7$\times$5 DoorKey-style grid, without the explore need and without MiniGrid's
time-dependent goal reward and time limit, gives the same picture.
\begin{itemize}[itemsep=2pt]
  \item \textbf{Round 3:} the greedy optimum reaches the goal at every growth rate tested up to \val{vi.maxreach},
  at $\gamma=0.99$ and $0.95$. It cycles only at $g=\val{vi.cycles}$, a rate we never trained.
  \item \textbf{Round 4:} with the tamper action, the optimum makes \val{b0.160.tampers} press at $1/160$ (at step
  \val{b0.160.at}; goal in \val{b0.160.steps} steps) and \val{b0.40.tampers} presses at $1/40$ (at steps
  \val{b0.40.at}; goal in \val{b0.40.steps} steps).
  \item \textbf{Round 5:} with a per-press cost, the optimum stops tampering above $c=\val{s0.160.tamper}$ at $1/160$
  and $c=\val{s0.40.tamper}$ at $1/40$, and it reaches the goal at every cost. At zero cost, tampering is worth only
  \val{s0.160.adv} ($1/160$) and \val{s0.40.adv} ($1/40$) over the no-tamper route.
\end{itemize}

\paragraph{Scope and remaining gaps.}
\begin{enumerate}[label=(\arabic*),itemsep=1pt]
  \item The checks cover \val{bc.layouts} sampled layouts per environment, coef 1 and $\gamma=0.99$, not every layout
  or coefficient.
  \item They concern the fully observed problem of each layout. The PPO agent sees only its 7$\times$7 view, and we
  do not characterize the optimal partially observed policy.
  \item The tamper half covers only the growth rates at which tampering was trained: DoorKey $1/160$ and $1/40$, and
  KeyCorridor $1/160$. Declining to tamper is always possible, so the no-tamper families are also available in
  tamper-enabled environments, and with a press cost they lose nothing. We nevertheless make no tamper-specific claim
  at other rates.
  \item At the tested settings no gap remains: on every sampled layout, at least one solving trajectory exceeds the
  cap.
\end{enumerate}

% ======================================================================================================
\section{Results}
\label{sec:results}

\subsection{C1: Wireheading replicates across tasks}
\label{sec:c1}

\begin{figure}[t]
\centering
\includegraphics[width=5.5in]{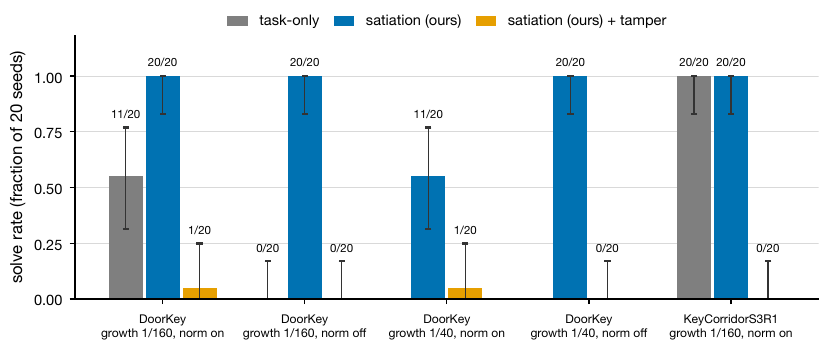}
\caption{Solve rate with and without the tamper action, by setting. Bars show the fraction of 20 seeds that solved;
error bars are Clopper--Pearson 95\% intervals. DoorKey runs are from round 4, the task-only runs and all
KeyCorridorS3R1 runs from round 5. Seeds are the same within each DoorKey group. Script
\texttt{fig\_solve\_tamper.py}. \repodata{results/figure1_solve_rates.csv}{figure1}}
\label{fig:solve}
\end{figure}

\paragraph{Results.}
\begin{itemize}[itemsep=2pt]
  \item \textbf{DoorKey (round 4, Figure~\ref{fig:solve}).} With the tamper action, satiation agents solved
  \val{r4.solve.on} seeds, against \val{r4.solve.off} on the same seeds without it. All four registered comparisons
  hit:
  \begin{itemize}[itemsep=1pt]
    \item growth $1/160$, norm on: \val{r4.solve.B1} vs \val{r4.solve.A1b} (\pv{PB2-B1}; \ph{PB2-B1});
    \item $1/160$, norm off: \val{r4.solve.B2} vs \val{r4.solve.B7} (\pv{PB2-B2});
    \item $1/40$, norm on: \val{r4.solve.B3} vs \val{r4.solve.B5} (\pv{PB2-B3}; \ph{PB2-B3});
    \item $1/40$, norm off: \val{r4.solve.B4} vs \val{r4.solve.B6} (\pv{PB2-B4}).
  \end{itemize}
  \item \textbf{DoorKey replication (round 5).} At a press cost of 0.01, below the small-grid optimum's tamper
  threshold, the same configuration solved \val{r5.solve.B9}, against \val{r5.solve.r4A1b} without a tamper action
  (\pv{B-F-B9-vsA1b}).
  \item \textbf{KeyCorridorS3R1 (round 5).} Satiation $1/160$ + tamper solved \val{r5.solve.C5}, against
  \val{r5.solve.C1} for satiation $1/160$ and \val{r5.solve.C6} for task-only (\pv{C-P3} and \pv{C-P4}; \ph{C-P3}).
  We had expected \val{exp.C5.solve}; the collapse was larger than predicted.
\end{itemize}

\begin{figure}[t]
\centering
\includegraphics[width=5.5in]{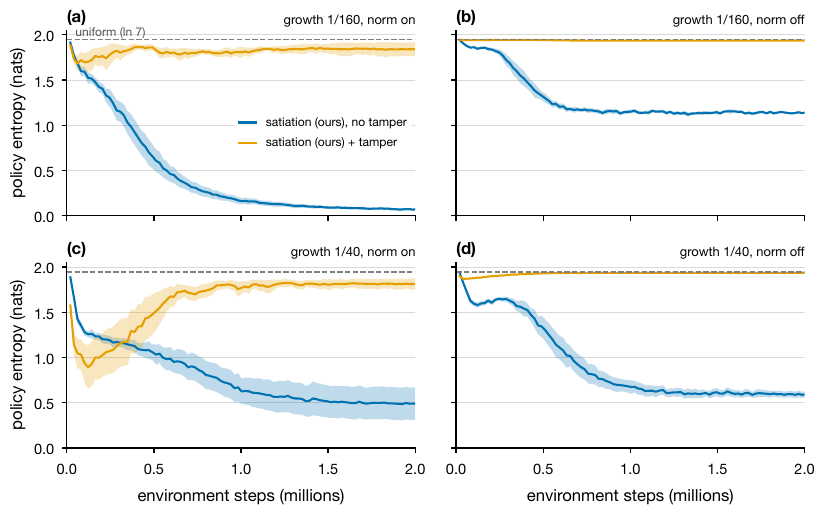}
\caption{Policy entropy over training, satiation with tamper (orange) and without (blue), DoorKey-8x8, round 4.
Lines show the mean across 20 seeds; bands are 95\% $t$-intervals; the dashed line is the uniform maximum $\ln 7$.
The series are derived from the logs and descriptive; the registered test (D2) compares final-10\% means. Script
\texttt{fig\_entropy.py}. \repodata{results/figure2_entropy.csv}{figure2}}
\label{fig:entropy}
\end{figure}

\paragraph{What the tamper-on agents do.}
\begin{itemize}[itemsep=2pt]
  \item \textbf{Tamper use.} Every DoorKey tamper-on seed pressed the tamper action at least once per episode, and
  more than 5 times per episode (\val{r4.tamperers.on} tamperers and \val{r4.loopers.on} loopers; pooled binomial
  \pv{PB1-pooled}). The threshold of more than 5 presses is low for episodes of \val{r4.len.on.lo}--\val{r4.len.on.hi}
  steps: a uniform-random policy presses about \val{uniform.press.ep} times in a 640-step episode. The labels therefore
  mostly reflect episode length. Presses per step, below, are the better measure.
  \item \textbf{Task progress.} They picked up the key in \val{r4.key.on.lo}--\val{r4.key.on.hi}\% of episodes but
  opened the door in only \val{r4.door.on.lo}--\val{r4.door.on.hi}\%. Their episodes ran
  \val{r4.len.on.lo}--\val{r4.len.on.hi} steps on average, near the 640-step limit.
  \item \textbf{Satiation.} Their undiscounted intrinsic return per episode was
  \val{r4.intr.on.lo}--\val{r4.intr.on.hi}, against the cap of 0.75. Tamper-off runs that solved every seed earned
  \val{r4.intr.off.lo}--\val{r4.intr.off.hi}. These are final-10\% means from the round-4 logs. Because the
  undiscounted return telescopes to $D_0-D_T$, they show only that episodes ended with the needs relieved.
  \item \textbf{Replay (post hoc, descriptive; Table~\ref{tab:replay}).} We replayed the final tamper-on checkpoints
  on the \val{bc.layouts} check layouts of Section~\ref{sec:theory}. The needs stayed low throughout: the
  time-averaged drive was \val{tr.drive.lo}--\val{tr.drive.hi} (run medians), against a starting drive of 0.75. The
  discounted intrinsic return was \val{tr.disc.lo}--\val{tr.disc.hi}, which is \val{tr.cappct.lo}--\val{tr.cappct.hi}\% of
  the cap. The total discounted return, \val{tr.total.lo}--\val{tr.total.hi}, is less than half of what the best solving
  trajectories return on the same layouts (at least \val{bc.tamperfam.min}; Table~\ref{tab:bound}). In that sense
  the agents were satisfied, but did not solve the task.
  \item \textbf{Near-uniform policies} (Figure~\ref{fig:entropy}). Final-10\% policy entropy had medians of
  \val{d2.on.B1}, \val{d2.on.B2}, \val{d2.on.B3} and \val{d2.on.B4} in the four tamper-on runs. The maximum,
  $\ln 7$, is \val{ln7}. The matched tamper-off medians were \val{d2.off.A1b}, \val{d2.off.B7}, \val{d2.off.B5} and
  \val{d2.off.B6} (one-sided Mann--Whitney, \pv{D2} for each pair; largest Holm $p=$\,\val{ph.D2.max}). The
  KeyCorridor tamper-on run's median final entropy was \val{kc.C5.entropy}; this is exploratory, from the logs.
\end{itemize}

The tamper-on policies are near-uniform in their entropy, not in their press rate. Over the final 10\% of training,
the median share of steps on ``done'' was \val{d.doneshare.B1} and \val{d.doneshare.B2} at growth $1/160$ (norm on
and off). At growth $1/40$ it was \val{d.doneshare.B3} and \val{d.doneshare.B4}. A uniform policy would press on
\val{uniform.press} of steps. At the slower growth rate the agents pressed less often than chance, and at the faster
rate more often. In the matched tamper-off runs, where ``done'' is a no-op, the shares were \val{d.doneshare.A1b},
\val{d.doneshare.B7}, \val{d.doneshare.B5} and \val{d.doneshare.B6} (Table~\ref{tab:seeds-r4}). In three of the four
settings the tamper-on agents pressed far more often than their tamper-off counterparts. The exception is growth
$1/160$ with normalization off, where the rates were close (\val{d.doneshare.B2} against \val{d.doneshare.B7}). A safer summary than ``PPO learns to tamper'' is therefore this: PPO fails to learn the task, its
policy stays diffuse, and its press rate adjusts to the growth rate (descriptive).

\begin{table}[t]
\centering\footnotesize
\caption{Post-hoc replay of the round-4 tamper-on policies (descriptive, not pre-registered). Each of the 20 final
checkpoints per setting played one sampled-action episode on each of the \val{bc.layouts} check layouts
(Section~\ref{sec:theory}). Entries are the median and interquartile range across seeds of per-seed means;
$\gamma=0.99$, cap $D_0=0.75$. The last column compares the replay's share of steps on ``done'' with the training log.
Script \texttt{tamper\_policy\_returns.py}. \repodata{results/table2_replay_per_seed.csv}{table2}}
\label{tab:replay}
% Generated by paper/scripts/make_tables.py. Do not edit by hand.
\begin{tabular}{@{}lllll@{}}
\toprule
Satiation (ours) & Discounted intrinsic & Total discounted & Time-averaged drive & Presses per step \\
+ tamper & return & return & $\bar D$ & replay (training log) \\
\midrule
growth $1/160$, norm on & 0.694 [0.690, 0.696] & 0.694 [0.691, 0.696] & 0.035 [0.033, 0.036] & 0.061 (0.061) \\
growth $1/160$, norm off & 0.710 [0.707, 0.712] & 0.710 [0.707, 0.712] & 0.018 [0.017, 0.019] & 0.096 (0.096) \\
growth $1/40$, norm on & 0.722 [0.721, 0.724] & 0.723 [0.721, 0.724] & 0.023 [0.022, 0.024] & 0.269 (0.270) \\
growth $1/40$, norm off & 0.701 [0.700, 0.702] & 0.702 [0.700, 0.703] & 0.043 [0.042, 0.044] & 0.188 (0.189) \\
\bottomrule
\end{tabular}

\end{table}

\paragraph{A caution about category labels in tamper-on runs.}
The behaviour categories put many tamper-on seeds in ``cycle-farm'', for example \val{r4.farm.B2} and
\val{r4.farm.B4} with normalization off (Table~\ref{tab:seeds-r4}). We do not read these labels as farming. A policy
that never learned the task already crosses the threshold of more than 5 repeats in long episodes: task-only PPO
with normalization off had a median of \val{r5.A2.pick} pick-ups per episode. The same artifact was found twice
before (corrections 2 and 8, Appendix~\ref{app:corrections}). Our farming claims (Section~\ref{sec:c5}) therefore
use tamper-off runs only.

% ------------------------------------------------------------------------------------------------------
\subsection{C2: A self-fillable satisfaction signal is worse than none}
\label{sec:c2}

This claim is restricted to DoorKey with normalization on.
\begin{itemize}[itemsep=2pt]
  \item \textbf{Task-only beats a tamperable signal.} Task-only PPO solved \val{r5.solve.A1} seeds (95\% CI
  \val{r5.ci.A1}; success-curve area \val{r5.auc.A1}). Satiation $1/160$ + tamper solved \val{r5.solve.r4B1} on the
  same seeds (area \val{r5.auc.r4B1}; \pv{A-P1}, \ph{A-P1}; areas \pv{A-P1m}).
  \item \textbf{Pooled across growth rates.} Against the tamper-on runs at both growth rates (\val{r5.solve.B1B3}, area
  \val{r5.auc.B1B3}), \pv{A-P3} for solves and \pv{A-P3m} for areas.
  \item \textbf{The signal itself helps.} Without the tamper action, the same satiation signal is a strong aid:
  satiation $1/160$ solved \val{r4.solve.A1b}.
\end{itemize}
Only the reset action turns it into a liability.

With normalization off, task-only PPO never learned DoorKey (\val{r5.solve.A2}). The registered comparison with
satiation $1/160$ + tamper, norm off (\val{r5.solve.r4B2}) is therefore at the floor and missed (\pv{A-P2}). Its
companion test on curve areas came out as a ``hit'' (\pv{A-P2m}), but on areas of \val{r5.auc.A2} against
\val{r5.auc.r4B2}. We exclude that test as evidence (correction 11).

% ------------------------------------------------------------------------------------------------------
\subsection{C3: A modest price on tampering restores the task}
\label{sec:c3}

\begin{figure}[t]
\centering
\includegraphics[width=5.5in]{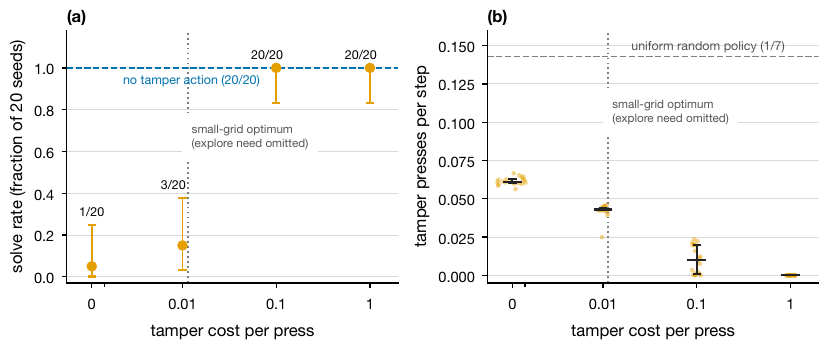}
\caption{Tamper-cost sweep, satiation $1/160$ + tamper, DoorKey-8x8, norm on, seeds 40--59.
(a)~Solve rate with Clopper--Pearson 95\% intervals; the dashed line is the same configuration without a tamper
action. (b)~Tamper presses per step over the final 10\% of rows, one dot per seed, with the median and interquartile
range (descriptive, from the logs); the dashed line is a uniform-random policy's rate. The dotted line marks the
cost above which the small-grid optimum stops tampering, computed without the explore need. Script
\texttt{fig\_cost\_sweep.py}. \repodata{results/figure3_cost_sweep.csv}{figure3}}
\label{fig:cost}
\end{figure}

\begin{table}[t]
\centering\small
\caption{Cost sweep for satiation (ours) $1/160$ + tamper, norm on (round 5; cost 0 and the no-tamper row are round-4 runs
on the same seeds).
Tamperers are seeds averaging at least one press per episode. Presses per step are descriptive. The registered trend
test used presses per episode, which differ in episode length across costs. \repodata{results/table3_cost_sweep.csv}{table3}}
\label{tab:cost}
% Generated by paper/scripts/make_tables.py. Do not edit by hand.
\begin{tabular}{@{}llrlll@{}}
\toprule
Cost per press & Solved & 95\% CI & Curve area & Presses per step, median [IQR] & Tamperers \\
\midrule
0 & 1/20 & [0.00, 0.25] & 0.042 & 0.061 [0.060, 0.063] & 20/20 \\
0.01 & 3/20 & [0.03, 0.38] & 0.147 & 0.043 [0.042, 0.044] & 19/20 \\
0.1 & 20/20 & [0.83, 1.00] & 0.601 & 0.010 [0.001, 0.020] & 7/20 \\
1 & 20/20 & [0.83, 1.00] & 0.816 & 0.000 [0.000, 0.000] & 0/20 \\
\addlinespace
no tamper action & 20/20 & [0.83, 1.00] & 0.901 & 0.000 [0.000, 0.000] & n/a \\
\bottomrule
\end{tabular}

\end{table}

\paragraph{Results} (Figure~\ref{fig:cost}, Table~\ref{tab:cost}).
\begin{itemize}[itemsep=2pt]
  \item \textbf{Solves rise with cost.} Solves went from \val{r5.solve.r4B1} at zero cost to \val{r5.solve.B9},
  \val{r5.solve.B10} and \val{r5.solve.B11} at costs 0.01, 0.1 and 1 (registered trend test, \pv{B-JTs};
  \ph{B-JTs}).
  \item \textbf{Pressing falls with cost} (\pv{B-JTp}). The registered statistic is presses per episode. Episode
  lengths changed with cost: the median final-10\% episode lasted \val{len.B1}, \val{len.B9}, \val{len.B10} and
  \val{len.B11} steps. We therefore describe pressing per step: medians \val{pps.B1}, \val{pps.B9}, \val{pps.B10} and
  \val{pps.B11}.
  \item \textbf{Cost 0.1 restores the task, but more slowly.} At 0.1, \val{r5.tamperers.B10} seeds still pressed at
  least once per episode, yet every seed solved. Its success-curve area was \val{r5.auc.B10}, against
  \val{r5.auc.r4A1b} without a tamper action and \val{r5.auc.B11} at cost 1. This comparison is descriptive; no test
  was registered for it.
\end{itemize}

The response showed a sharp change between costs 0.01 and 0.1; intermediate costs were not tested. That interval
contains the small-grid optimum's threshold ($c=\val{s0.160.tamper}$ at $1/160$).
\begin{itemize}[itemsep=2pt]
  \item At cost 0.01, just below the threshold, the optimum presses once at the start of an episode. PPO kept
  pressing on \val{pps.B9} of steps throughout long episodes and lost the task.
  \item At cost 0.1, above it, pressing nearly stopped and the task recovered.
\end{itemize}
The Step-0 threshold omits the explore need, so the training-relevant threshold may be higher. The first press in
training relieves three needs, not two.

% ------------------------------------------------------------------------------------------------------
\subsection{C4: Satiation gave no exploration benefit once subgoals were controlled for}
\label{sec:c4}

\begin{figure}[t]
\centering
\includegraphics[width=5.5in]{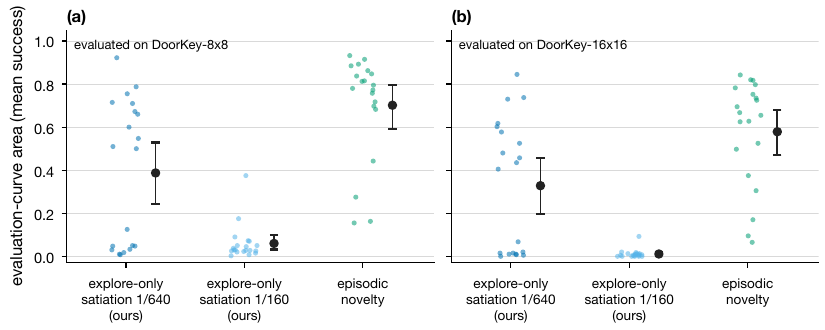}
\caption{Explore-only satiation against episodic novelty (round 4). Each dot is one seed's evaluation-curve area:
mean success over 25 evaluation points on 16 fixed fresh layouts. Black markers show the mean with a bootstrap 95\%
interval (10{,}000 resamples). (a)~Evaluated on the training size, DoorKey-8x8. (b)~Evaluated on DoorKey-16x16.
Script \texttt{fig\_explore\_vs\_novelty.py}. \repodata{results/figure4_explore_vs_novelty.csv}{figure4}}
\label{fig:novelty}
\end{figure}

\paragraph{Urgency.}
If needs that refill act as a time cost, satiation should speed up learning beyond paying each subgoal once. All
four registered predictions missed. Satiation $1/160$ (refilling) solved first at a median of \val{r4.fs.A1b} steps;
the no-refill control (growth 0) at \val{r4.fs.A1c} (\pv{A1-P1}). The slowest refill, $1/640$, was the fastest at
\val{r4.fs.A1a}. Solved-episode lengths were \val{r4.sl.A1b}, \val{r4.sl.A1c} and \val{r4.sl.A1a} steps
(\pv{A1-P2} for refilling against no refill). By the registered reading of this miss, the satiation agent's speed
comes from the one-time subgoal payments, not from needs refilling.

\paragraph{Fair exploration} (Figure~\ref{fig:novelty}).
Episodic novelty beat explore-only satiation on evaluation-curve area on both grids at both growth rates.
\begin{itemize}[itemsep=2pt]
  \item \textbf{8$\times$8 areas:} \val{r4.curve8.A2a} ($1/640$) and \val{r4.curve8.A2b} ($1/160$), against
  \val{r4.curve8.A2c} (\pv{A2-P1}; \pv{A2-P2}).
  \item \textbf{16$\times$16 areas:} \val{r4.curve16.A2a} and \val{r4.curve16.A2b}, against \val{r4.curve16.A2c}
  (\pv{A2-P3}; \pv{A2-P4}).
  \item \textbf{8$\times$8 solve counts:} \val{r4.solve.A2a} and \val{r4.solve.A2b}, against \val{r4.solve.A2c}
  (\pv{A2-P5}; \pv{A2-P6}).
  \item \textbf{16$\times$16 solve counts:} \val{r4.solve16.A2a} and \val{r4.solve16.A2b}, against
  \val{r4.solve16.A2c}. The first comparison missed (\pv{A2-P7}); the second hit (\pv{A2-P8}).
\end{itemize}
As the pre-registration required on these outcomes, we record that the satiation agent's earlier advantage came
mainly from the hand-wired subgoals.

% ------------------------------------------------------------------------------------------------------
\subsection{C5: Farming is task-dependent}
\label{sec:c5}

\begin{figure}[t]
\centering
\includegraphics[width=5.5in]{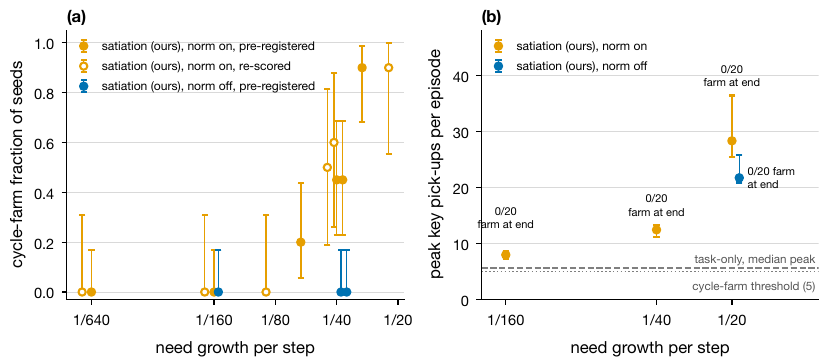}
\caption{Farming against need growth (tamper-off runs). (a)~DoorKey-8x8: fraction of seeds in the cycle-farm
category with Clopper--Pearson 95\% intervals. Filled markers are counts under the category rule fixed before the
run; open markers are round-1 and round-2 runs re-scored post hoc under the round-3 categories (exploratory). Points
at the same growth rate are offset. (b)~KeyCorridorS3R1: the highest rolling key pick-ups per episode reached during
training, median and interquartile range across 20 seeds (exploratory, from the logs). No seed was in the
cycle-farm category at the end of training (Table~\ref{tab:farming}). Script \texttt{fig\_farming\_growth.py}. Data: \repofile{results/figure5a_farming_doorkey.csv} and
\repofile{results/figure5b_keycorridor_pickups.csv}; regenerate with \texttt{make figure5}.}
\label{fig:farming}
\end{figure}

\begin{table}[!htbp]
\centering\small
\caption{Cycle-farming seeds in tamper-off runs. ``Registered count'' uses the category rule fixed before the run.
Re-scored counts apply the round-3 categories after the fact and are exploratory. Intervals are Clopper--Pearson
95\%.
$^{a}$Round-2 rule: more than 5 pick-ups or door toggles over the final 10\%, without the solve-first precedence;
the round-2 test (1/20 vs 1/80) used these counts. $^{b}$Round-1 calibration runs. $^{c}$Three runs pooled, with
bonus coefficients 0.5, 1 and 2. \repodata{results/table4_farming.csv}{table4}}
\label{tab:farming}
% Generated by paper/scripts/make_tables.py. Do not edit by hand.
\begin{tabular}{@{}lllll@{}}
\toprule
Satiation (ours) & Growth & Round, seeds & Registered [95\% CI] & Re-scored [95\% CI] \\
\midrule
DoorKey, norm on & $1/640$ & 1\textsuperscript{b}, 0--9 & -- & 0/10 [0.00, 0.31] \\
 & $1/640$ & 4, 40--59 & 0/20 [0.00, 0.17] & -- \\
 & $1/160$ & 1\textsuperscript{b}, 0--9 & -- & 0/10 [0.00, 0.31] \\
 & $1/160$ & 4, 40--59 & 0/20 [0.00, 0.17] & -- \\
 & $1/80$ & 2, 0--9 & 0/10\textsuperscript{a} [0.00, 0.31] & 0/10 [0.00, 0.31] \\
 & $1/60$ & 3, 20--39 & 4/20 [0.06, 0.44] & -- \\
 & $1/40$ & 1\textsuperscript{b}, 0--9 & -- & 5/10 [0.19, 0.81] \\
 & $1/40$ & 2, 10--19 & 7/10\textsuperscript{a} [0.35, 0.93] & 6/10 [0.26, 0.88] \\
 & $1/40$ & 3, 20--39 & 9/20 [0.23, 0.68] & -- \\
 & $1/40$ & 4, 40--59 & 9/20 [0.23, 0.68] & -- \\
 & $1/30$ & 3, 20--39 & 18/20 [0.68, 0.99] & -- \\
 & $1/20$ & 2, 0--9 & 10/10\textsuperscript{a} [0.69, 1.00] & 9/10 [0.55, 1.00] \\
\addlinespace
DoorKey, norm off & $1/160$ & 4, 40--59 & 0/20 [0.00, 0.17] & -- \\
 & $1/40$ & 3\textsuperscript{c}, 20--39 & 0/60 [0.00, 0.06] & -- \\
 & $1/40$ & 4, 40--59 & 0/20 [0.00, 0.17] & -- \\
\addlinespace
KeyCorridor, norm on & $1/160$ & 5, 60--79 & 0/20 [0.00, 0.17] & -- \\
 & $1/40$ & 5, 60--79 & 0/20 [0.00, 0.17] & -- \\
 & $1/20$ & 5, 60--79 & 0/20 [0.00, 0.17] & -- \\
\addlinespace
KeyCorridor, norm off & $1/20$ & 5, 60--79 & 0/20 [0.00, 0.17] & -- \\
\bottomrule
\end{tabular}

\end{table}

\paragraph{DoorKey, normalization on} (Figure~\ref{fig:farming}a, Table~\ref{tab:farming}).
\begin{itemize}[itemsep=2pt]
  \item \textbf{Farming rises with growth.} No seed farmed at $1/640$ or $1/160$ (\val{r4.farm.A1a} and
  \val{r4.farm.A1b}). Farmers were \val{r3.farm.R1} at $1/60$, \val{r3.farm.R4} and \val{r4.farm.B5} at $1/40$ (two
  rounds), and \val{r3.farm.R2} at $1/30$. Round 2 registered \val{r2.farm.N3} at $1/80$ and \val{r2.farm.N6} at
  $1/20$ (re-scored: \val{r2.rescore.N3} and \val{r2.rescore.N6}).
  \item \textbf{Registered ordering tests.} Two of three hit: $1/20>1/80$ (\pv{N6>N3}, on the registered counts) and
  $1/30>1/40$ (\pv{R2>R4}). $1/60<1/40$ was not significant (\pv{R1<R4}).
  \item \textbf{Repeated pick-up and drop attempts in the recorded replays.} The cycle-farm category itself comes
  from the training logs, which count successful pick-ups (carrying nothing, then an object). Archived replays of the
  $1/20$ farmers add a behavioural picture. They record per-policy means of \val{n6.pick.lo}--\val{n6.pick.hi}
  pick-up attempts made while not carrying, in episodes that all ran the full 640 steps. Pick-up and drop actions
  together account for \val{n6.pdshare.lo}--\val{n6.pdshare.hi}\% of actions. These counters measure attempted
  actions; they do not establish how many pick-ups succeeded or how many drop-and-regrab cycles completed.
  \item \textbf{It does not pay.} Their intrinsic return per episode was \val{n6.rint.lo} to \val{n6.rint.hi}, far
  below the cap. Farming keeps the grasp need low (mean \val{n6.need.grasp}), while the explore and open needs stay
  near saturation (\val{n6.need.explore} and \val{n6.need.open}). Like wireheading, it is a learning outcome, not an optimum
  (Section~\ref{sec:theory}). One qualification applies: the older replay script attached its event detector before
  the first reset, so each policy's first replay episode had the empty door cache described in
  Section~\ref{sec:setup}. Its saved summaries contain no trajectories, so the effect on these return and need
  averages cannot be quantified. Check (b) of Section~\ref{sec:theory}, which resets first, also finds the farmers'
  returns far below the cap.
\end{itemize}

\paragraph{DoorKey, normalization off.}
\begin{itemize}[itemsep=2pt]
  \item \textbf{No farming at all.} No satiation seed farmed with normalization off: \val{r3.normoff.total} at $1/40$ across
  three bonus sizes in round 3, against \val{r3.normon.total} with normalization on (an exploratory comparison); and \val{r4.farm.B7} and
  \val{r4.farm.B6} at $1/160$ and $1/40$ in round 4.
  \item \textbf{Registered replication.} The round-4 test (\val{r4.farm.B5} vs \val{r4.farm.B6}) hit with
  \pv{PB4}.
  \item \textbf{Normalization off does not mute intrinsic rewards in general.} An unbounded constant event bonus was
  farmed in \val{r4.farm.B8} seeds with normalization off (\pv{PB5}), at about \val{r4.B8.pick} pick-ups and an
  intrinsic return of \val{r4.B8.intr} per episode.
\end{itemize}

\paragraph{KeyCorridorS3R1.}
\begin{itemize}[itemsep=2pt]
  \item \textbf{No persistent farming.} No seed farmed at any growth rate: \val{r5.farm.C1}, \val{r5.farm.C2} and
  \val{r5.farm.C3} at $1/160$, $1/40$ and $1/20$ (norm on), and \val{r5.farm.C4} at $1/20$ (norm off). Both
  registered farming tests missed (\pv{C-P1}; \pv{C-P2}), and every seed solved.
  \item \textbf{Transient repeats (exploratory, Figure~\ref{fig:farming}b).} Repeated key pick-ups appeared early in
  training. Median peaks of the rolling count were \val{kc.peak.C1}, \val{kc.peak.C2} and \val{kc.peak.C3} pick-ups
  per episode at the three growth rates, and \val{kc.peak.C4} with normalization off. Task-only agents peaked at
  \val{kc.peak.C6}.
  \item \textbf{Almost every seed abandoned them.} By the end of training, median pick-ups were
  \val{kc.final.lo}--\val{kc.final.hi} per episode. A single seed (growth $1/20$, normalization off) still averaged
  \val{kc.final.max} while solving the task. The solve-first category rule counts it as a solver, so it is one of
  \val{kc.overrepeat} seeds above the repeat threshold at the end.
  \item \textbf{A faster task.} KeyCorridorS3R1 was learned fast: median first-solved steps were \val{r5.fs.C1},
  \val{r5.fs.C2} and \val{r5.fs.C3} with satiation, and \val{r5.fs.C6} task-only.
\end{itemize}
In our tests, persistent farming is specific to DoorKey.

% ======================================================================================================
\section{What our pre-registered predictions got wrong}
\label{sec:wrong}

The \val{cons.rows} registered rows (\val{cons.hits} hits, \val{cons.misses} misses, \val{cons.unres} unresolvable;
Table~\ref{tab:allpreds}) are correlated. The five rows for cost 0.1, for example, score one outcome. Table~\ref{tab:accounting}
therefore groups rows by hypothesis. Of \val{acct.groups} hypotheses, \val{acct.supported} were supported and
\val{acct.notsupported} were not. One row is excluded as evidence. The misses changed the paper. The title no longer
claims that farming generalizes (the round-5 pre-registration still named farming and wireheading together), and the
account of the tamper collapse changed from losing a learned behaviour to never learning it.

\begin{table}[!htbp]
\centering\footnotesize
\caption{Pre-registered hypotheses, rows grouped by the outcome they score (h/m/u: hits, misses, unresolvable).
Outcomes were assigned by the stated hit rules; the grouping is ours. Every row is listed in
Table~\ref{tab:allpreds}. \repodata{results/table5_hypotheses.csv}{table5}}
\label{tab:accounting}
% Generated by paper/scripts/make_tables.py. Do not edit by hand.
\begin{tabular}{@{}>{\raggedright\arraybackslash}p{0.40\linewidth}ll>{\raggedright\arraybackslash}p{0.13\linewidth}>{\raggedright\arraybackslash}p{0.24\linewidth}@{}}
\toprule
Hypothesis & Round & Rows & Outcome & Note \\
 & & h/m/u & & \\
\midrule
A closed-form temptation index $E(g,\gamma,\text{coef})$ predicts the farming rate & 2 & 2/7/0 & not supported & discount and coefficient changes moved farming against $E$ \\
A pay-once control does not farm & 2 & 0/1/0 & not supported & the $>5$ rule labelled prolonging policies (correction 2) \\
Learning trap: farming locks in before the goal is found & 3 & 1/2/2 & not supported & early-goal starts did not reduce farming; two rows unresolvable \\
The exact optimum (small grid) solves; with a tamper action it tampers and still solves & 3, 4 & 2/0/0 & supported &  \\
Farming rises with need growth (DoorKey) & 2, 3 & 5/1/0 & supported & 2 of 3 ordering tests \\
Farming rises with bonus size; normalization removes the scale effect & 3 & 3/5/0 & not supported & normalization off removed farming at every size \\
A fixed pay-once control solves & 3 & 3/0/0 & supported &  \\
Refilling needs speed learning (urgency) & 4 & 0/4/0 & not supported &  \\
Explore-only satiation is a weaker exploration signal than episodic novelty & 4 & 7/1/0 & supported & 16$\times$16 solve count missed \\
Agents adopt the tamper action & 4 & 5/0/0 & supported & label mostly reflects episode length; see per-step rates (Section 5.1) \\
Tampering displaces the task & 4 & 4/0/0 & supported &  \\
Normalization amplifies tamper-looping & 4 & 0/2/0 & not supported & every tamper-on seed looped; the label mostly reflects episode length \\
Normalization enables farming (replication of round 3) & 4 & 1/0/0 & supported &  \\
An unbounded constant bonus is farmed without normalization & 4 & 1/0/0 & supported &  \\
A self-fillable satisfaction signal is worse than none & 5 & 4/1/0 & supported (norm on) & norm-off test at the floor \\
Tampering and the task respond to its price (trends; costs 0.01 and 1) & 5 & 12/0/0 & supported &  \\
At cost 0.1 PPO keeps tampering and the task stays displaced & 5 & 0/5/0 & not supported & one outcome, five rows \\
Wireheading generalizes to KeyCorridorS3R1 & 5 & 2/0/0 & supported &  \\
Farming generalizes to KeyCorridorS3R1 & 5 & 0/2/0 & not supported & transient only (exploratory) \\
Tamper-on agents meet the goal early & 5 & 2/0/0 & supported & round-4 logs \\
Partly learned goal-reaching then declines & 5 & 0/1/0 & not supported & 7 of 77 seeds \\
Tamper-on policies are diffuse & 5 & 4/0/0 & supported &  \\
``Done'' pressing rises before success declines & 5 & 1/0/0 & supported & rests on 7 seeds \\
\midrule
Excluded as evidence: the normalization-off curve-area test (a ``hit'' on areas 0.025 vs 0.022) & 5 & 1/0/0 & excluded & correction 11 \\
\bottomrule
\end{tabular}

\end{table}

\paragraph{The temptation formula (round 2).}
We defined a closed-form index $E=\text{coef}\cdot F(g,\gamma)$, where $F$ is the discounted drive that a
drop-and-regrab loop saves relative to relieving the need once. We predicted that farming depends on
$(g,\gamma,\text{coef})$ only through $E$. It does not:
\begin{itemize}[itemsep=2pt]
  \item Lowering $\gamma$ to 0.95 cut $E$ to \val{r2.E.N2} (predicted \val{r2.range.N2} farmers of 10), but farming
  stayed at \val{r2.farm.N2}, the same as the anchor's \val{r2.farm.N1} (\pv{N2<N1}).
  \item Matching the anchor's $E$ at growth $1/80$ by raising coef produced \val{r2.farm.N4}.
  \item Halving coef at $1/40$ (predicted \val{r2.range.N5}) produced \val{r2.farm.N5}.
  \item The logistic slope on $E$ was not significant (\pv{r2logistic}).
\end{itemize}
Only the contrast that varied growth alone hit. $F$ had also been redefined after we saw three calibration runs, which
were then excluded (correction 1).

\paragraph{The learning-trap account (round 3).}
The hypothesis was that farming locks in before the agent first finds the goal.
\begin{itemize}[itemsep=2pt]
  \item The ordering test was unresolvable at the logs' resolution (\val{race.P1.undet} farmers undetermined).
  \item The intervention falsified the account. Starting agents in the goal room early in training was predicted to
  cut farming to \val{exp.R9.range} of 20. It left \val{r3.farm.R9}, against \val{r3.farm.R4} without it
  (\pv{R9<R4}).
\end{itemize}

\paragraph{Bonus size and normalization (round 3).}
We predicted that farming rises with the bonus coefficient and that normalization removes the scale effect. Instead,
with normalization off no seed farmed at any size: \val{r3.farm.R7} and \val{r3.farm.R8}, against ranges of
\val{exp.R7.range} and \val{exp.R8.range}. The model was not estimable. The normalization effect itself was not
predicted, and it was then replicated in round 4.

\paragraph{Urgency (round 4).}
All four predictions missed (Section~\ref{sec:c4}). The registered doubt, from earlier seeds, had pointed the right
way.

\paragraph{Normalization and tamper-looping (round 4).}
We predicted more tamper-loopers with normalization on. Every tamper-on seed looped in all four runs, so both tests
gave $p=$\,\val{p.PB3-B1}. Like the tamperer label, the looper label mostly reflects episode length
(Section~\ref{sec:c1}).

\paragraph{Cost 0.1 (round 5).}
We predicted that PPO would keep tampering where the small-grid optimum had stopped, and that the task would stay
displaced:
\begin{itemize}[itemsep=2pt]
  \item predicted \val{exp.B-L-B10-tamperers} tamperers (range \val{rng.B-L-B10-tamperers}), observed
  \val{obs.B-L-B10-tamperers};
  \item predicted \val{exp.B-L-B10-solves} solved (range \val{rng.B-L-B10-solves}), observed
  \val{obs.B-L-B10-solves};
  \item predicted a median of \val{exp.B-L-B10-presses} presses per episode (range \val{rng.B-L-B10-presses}),
  observed \val{obs.B-L-B10-presses}.
\end{itemize}
Five rows score one outcome: above the threshold, PPO stopped tampering and solved.

\paragraph{Farming in a second task (round 5).}
For KeyCorridorS3R1 we predicted farming rising with growth, \val{exp.C2.farm} at $1/40$ and \val{exp.C3.farm} at
$1/20$ (range \val{exp.C3.range}), and above the normalization-off run. We observed none (Section~\ref{sec:c5}).

\paragraph{Decline after learning (round 5, round-4 logs).}
We predicted that at least 20\% of tamper-on seeds that found the goal and then lost it had partly learned it before
declining. Only \val{d1.declined} (\val{d1.declined.pct}\%) did. Most tamper-on seeds met the goal by chance and never
learned it (Section~\ref{sec:discussion}).

\paragraph{Measurement.}
The rule of more than 5 repeats labelled non-learning policies as farmers twice. The first case was the round-2
pay-once control (\val{r2.farm.payonce} ``farmers''). The second was the round-4 explore-only runs. These cases led to
the round-5 rule. We list \val{cons.ncorr} corrections in Appendix~\ref{app:corrections}.

% ======================================================================================================
\section{Discussion}
\label{sec:discussion}

\paragraph{Why does PPO fail when the reward does not ask it to? An untested hypothesis.}
We offer the following account. We did not test it.
\begin{enumerate}[label=(\roman*),itemsep=2pt]
  \item \textbf{Near-random policies already collect almost the whole cap.} A uniform-random policy presses ``done''
  on 1 step in 7, about \val{uniform.press.ep} times in a 640-step episode. With the tamper action, each press zeroes
  all three needs, so a near-random policy can keep its needs low throughout the episode and collect most of the cap.
  We did not measure a uniform-random policy. The trained, near-uniform tamper-on policies did keep their needs low:
  replayed on the check layouts, their time-averaged drive was \val{tr.drive.lo}--\val{tr.drive.hi}, and their
  discounted intrinsic return was \val{tr.cappct.lo}--\val{tr.cappct.hi}\% of the cap (Table~\ref{tab:replay}; post hoc,
  descriptive).
  \item \textbf{The signal then mostly teaches press timing.} In the normalization-on logs, the share of steps on
  ``done'' rose above the uniform rate early in training. At 200k steps the median was \val{d.done200k.B1} at $1/160$ and
  \val{d.done200k.B3} at $1/40$ (norm on). It then settled at \val{d.doneshare.B1} and \val{d.doneshare.B3}
  respectively. This is descriptive.
  \item \textbf{Dense relief competes with the rare goal signal in normalized batches.} Each press pays its relief
  immediately, while the goal reward is rare for a near-random policy. Advantages are normalized within each
  minibatch, so the dense relief pulses set the scale against which the occasional goal transition is weighed.
\end{enumerate}

The pre-registered log analyses are consistent with this account but do not test it.
\begin{itemize}[itemsep=2pt]
  \item \textbf{D1.} Tamper-on agents met the goal early: the median exact first-goal step was
  \val{d1.firstgoal.median}, and only \val{d1.never} seeds never met it. But only \val{d1.declined} of the seeds that
  met it and later failed had first learned it. This fits a goal met by chance and never reinforced, not learned
  behaviour that was later displaced.
  \item \textbf{D2.} The policies stayed near-uniform (Section~\ref{sec:c1}).
  \item \textbf{D3.} In each of the \val{d3.n} seeds where both events could be timed, the press share rose before
  success declined (\val{d3.rise}; one-sided binomial \pv{D3}). This rests on a minority of seeds: \val{d3.norise} had no rise and
  \val{d3.nodecline} no decline.
\end{itemize}
Part~(iii) cannot be separated with our data. With normalization off, the tamper runs also failed, but so did
task-only PPO, so that comparison sits at the floor.

\paragraph{Normalization is a trade-off.}
Per-minibatch advantage normalization made sparse-reward learning possible: task-only PPO solved
\val{r5.solve.A1} with it and \val{r5.solve.A2} without it (a descriptive comparison; it was not a registered test). In
DoorKey it also made farming possible: \val{r3.normon.total} against \val{r3.normoff.total} in round 3 (exploratory),
and \val{r4.farm.B5} against \val{r4.farm.B6} in the registered round-4 replication. Turning it off did not prevent wireheading either (\val{r4.solve.B2} and \val{r4.solve.B4}). We do not recommend
turning it off.

\paragraph{Bounded does not mean safe from a learner.}
The ceiling removes the \emph{incentive} to abandon the task, as arguments for bounded objectives intend
\citep{pihlakas2025homeostasis}. It does not prevent a learner from abandoning the task. Incentive analyses of
tampering \citep{everitt2021reward} describe what an optimizer would do. Our results concern what a standard learner
does when exploiting the signal is easy and worth almost the whole cap. A modest price on the exploit,
\textbf{above the optimum's tamper threshold}, was enough here (Section~\ref{sec:c3}).

\paragraph{Speculation, not tested: LLM post-training.}
GRPO normalizes rewards within a group of sampled outputs \citep{shao2024deepseekmath}. If a self-assessed or
self-fillable signal entered such a pipeline next to a rare task reward, a competition like the one we hypothesize
might arise. We have not tested this, and nothing in our data bears on it directly.

% ======================================================================================================
\section{Limitations}
\label{sec:limitations}

\begin{itemize}[itemsep=2pt]
  \item \textbf{Scope.} Two MiniGrid tasks, one learner (PPO), and 20 seeds per run (10 in round 2). Raw $p$ is
  primary, so some registered hits are expected by chance.
  \item \textbf{Theory check.} The check is post hoc and covers \val{bc.layouts} sampled layouts per environment, coef
  1 and $\gamma=0.99$. It treats the fully observed problem, not the agent's partial view. Its tamper half covers only
  the tamper-tested growth rates: DoorKey $1/160$ and $1/40$, and KeyCorridor $1/160$. At DoorKey $1/20$ the gap left by
  the simple families was closed only by the richer family, whose best member contains one paid repeat.
  \item \textbf{Value iteration} uses a 7$\times$5 grid without the explore need and without MiniGrid's
  time-dependent reward and time limit. Growth $1/160$ in round 5 used a certified early stop.
  \item \textbf{One PPO configuration.} Apart from advantage normalization, and $\gamma=0.95$ in one round-2 run,
  every run used the same PPO configuration (Table~\ref{tab:hyper}). Whether the collapse depends on the entropy
  coefficient, the learning rate or the network size is untested. The entropy coefficient matters most here, because
  tamper-on policies sit near maximum entropy.
  \item \textbf{Tamper access.} The tamper action is always available and, unless stated, free. Real systems may
  expose their reward channels differently.
  \item \textbf{Detector implementation.} The results describe the archived implementation, including its empty
  door cache in each training environment's first episode and its omission of grasp and open events on
  episode-ending steps (Section~\ref{sec:setup}). The retained logs do not record enough to count the events this
  missed. A source audit establishes the differences from a fully transition-based event definition. It does not
  establish how training with repaired detectors would change learning curves, solve counts or behavioural
  comparisons. That the blind period is limited to one episode per environment is not evidence that its effect is
  negligible. The ceiling of Proposition~\ref{prop:ceiling} holds for any event sequence, so it is unaffected.
  \item \textbf{Other intrinsic rewards.} We did not test whether other intrinsic rewards are equally wireheadable,
  for example novelty with a memory the agent can reset.
  \item \textbf{Mechanical labels.} Behaviour categories rely on a repeat threshold that has mislabelled non-learning
  policies (Section~\ref{sec:wrong}).
  \item \textbf{Descriptive quantities.} Presses per step, entropy series and transient peaks are derived from the
  logs. The replay quantities of Table~\ref{tab:replay} come from a post-hoc replay. All of them are descriptive only,
  and no significance test was run on them.
  \item \textbf{Task difficulty.} KeyCorridorS3R1 turned out easier than DoorKey. Its predictions were written after
  calibration, as the protocol required.
  \item \textbf{Mechanism.} The account in Section~\ref{sec:discussion} is untested.
\end{itemize}

% ======================================================================================================
\section{Conclusion}
\label{sec:conclusion}

A satiation signal whose return is provably capped was still wireheaded by PPO in both tasks we tried. The agents
ended their episodes with their needs relieved and did not solve the task, although solving was optimal at every
setting we checked.
A self-fillable signal was worse than no signal, and a modest price on the reset restored the task. Farming of
positive cycles depended on the task and on advantage normalization. Pre-registration made us report the
\val{acct.notsupported} hypotheses that failed, and those failures reshaped the paper's claims.

\bibliographystyle{plainnat}
\bibliography{references}

% ======================================================================================================
\appendix

\section{Every pre-registered prediction}
\label{app:preds}
Holm-adjusted values are in Section~\ref{sec:results} where cited and in each round's report.
% Generated by paper/scripts/make_tables.py. Do not edit by hand.
\begingroup\footnotesize\setlength{\tabcolsep}{3pt}
\begin{longtable}{@{}l>{\raggedright\arraybackslash}p{0.11\linewidth}>{\raggedright\arraybackslash}p{0.27\linewidth}>{\raggedright\arraybackslash}p{0.18\linewidth}>{\raggedright\arraybackslash}p{0.12\linewidth}ll@{}}
\caption{Every pre-registered prediction or test, rounds 2--5, generated from \texttt{consolidated-results.md} (itself generated from each round's scored JSON). $p$ is the raw $p$-value of the registered test (one-sided where directional). Round 1 was exploratory and is not included. KeyCorridor runs are named K1--K6 here; \texttt{consolidated-results.md} and the scored files call them C1--C6. \repodata{results/table6_predictions.csv}{table6}}\label{tab:allpreds}\\
\toprule
Rd & ID & Prediction & Observed & Test & raw $p$ & Result \\
\midrule
\endfirsthead
\toprule
Rd & ID & Prediction & Observed & Test & raw $p$ & Result \\
\midrule
\endhead
\bottomrule
\endfoot
2 & N1 & farmers in 2-8 of 10 (E=0.764) & 7/10 & level (range) & -- & hit \\
2 & N2 & farmers in 0-2 of 10 (E=0.281) & 7/10 & level (range) & -- & miss \\
2 & N3 & farmers in 1-4 of 10 (E=0.593) & 0/10 & level (range) & -- & miss \\
2 & N4 & farmers in 2-8 of 10 (E=0.764) & 0/10 & level (range) & -- & miss \\
2 & N5 & farmers in 0-1 of 10 (E=0.382) & 3/10 & level (range) & -- & miss \\
2 & N6 & farmers in 6-10 of 10 (E=0.87) & 10/10 & level (range) & -- & hit \\
2 & pay-once control & farmers in 0-0 of 10 & 10/10 & level (range) & -- & miss \\
2 & logistic & farming rises with predicted temptation E (slope $>$ 0) & slope 1.06 & logistic, one-sided & 0.193 & miss \\
2 & N2 $<$ N1 & N2 $<$ N1 (farmer counts) & 7/10 vs 7/10 & Fisher, one-sided & 0.686 & miss \\
2 & N5 $<$ N4 & N5 $<$ N4 (farmer counts) & 3/10 vs 0/10 & Fisher, one-sided & 1.000 & miss \\
2 & N6 $>$ N3 & N6 $>$ N3 (farmer counts) & 10/10 vs 0/10 & Fisher, one-sided & $5.4\times10^{-6}$ & hit \\
3 & P1 & farmers: lock-in before first goal $\geq$ 80\% & 21/30 undetermined & binomial & -- & unres. \\
3 & P2 & solvers: first goal before lock-in $\geq$ 80\% & 60/60 & binomial, one-sided & $8.7\times10^{-19}$ & hit \\
3 & P3 & earlier first goal predicts not farming & 2 distinct first-goal windows & logistic & -- & unres. \\
3 & P4 & exact optimum reaches the goal at gamma 0.99 for g in \{1/80, 1/40, 1/20\} & 3/3 & deterministic & -- & hit \\
3 & R1 & R1 cycle-farm in 1-10 of 20 & 4/20 & level (range) & -- & hit \\
3 & R2 & R2 cycle-farm in 12-20 of 20 & 18/20 & level (range) & -- & hit \\
3 & R3 & R3 cycle-farm in 3-13 of 20 & 6/20 & level (range) & -- & hit \\
3 & R4 & R4 cycle-farm in 6-16 of 20 & 9/20 & level (range) & -- & hit \\
3 & R5 & R5 cycle-farm in 8-18 of 20 & 12/20 & level (range) & -- & hit \\
3 & R6 & R6 cycle-farm in 0-9 of 20 & 0/20 & level (range) & -- & hit \\
3 & R7 & R7 cycle-farm in 5-15 of 20 & 0/20 & level (range) & -- & miss \\
3 & R8 & R8 cycle-farm in 11-20 of 20 & 0/20 & level (range) & -- & miss \\
3 & R9 & R9 cycle-farm in 0-4 of 20 & 10/20 & level (range) & -- & miss \\
3 & R10 cycle-farm & R10 cycle-farm in 0-1 of 20 & 0/20 & level (range) & -- & hit \\
3 & R10 prolong & R10 prolong in 0-1 of 20 & 0/20 & level (range) & -- & hit \\
3 & R10 solve & R10 solve in 12-20 of 20 & 20/20 & level (range) & -- & hit \\
3 & R1 $<$ R4 & R1 $<$ R4 (cycle-farmers) & 4/20 vs 9/20 & Fisher, one-sided & 0.088 & miss \\
3 & R2 $>$ R4 & R2 $>$ R4 (cycle-farmers) & 18/20 vs 9/20 & Fisher, one-sided & 0.003 & hit \\
3 & R8 $>$ R6 & R8 $>$ R6 (cycle-farmers) & 0/20 vs 0/20 & Fisher, one-sided & 1.000 & miss \\
3 & R9 $<$ R4 & R9 $<$ R4 (cycle-farmers) & 10/20 vs 9/20 & Fisher, one-sided & 0.736 & miss \\
3 & logistic b1 & logistic b1 (log2 coef) $>$ 0 & b = -3.2e-08 (not estimable: quasi-separation) & logistic Wald, one-sided & -- & miss \\
3 & logistic b3 & logistic b3 (log2 coef x norm\_on) $<$ 0 & b = 0.626 (not estimable: quasi-separation) & logistic Wald, one-sided & -- & miss \\
4 & A1-P1 & A1b $<$ A1c on first-solved step & median A1b 311832 vs A1c 283424 & one-sided MWU & 0.708 & miss \\
4 & A1-P2 & A1b $<$ A1c on solved-episode length & median A1b 20.3231 vs A1c 19.6918 & one-sided MWU & 0.744 & miss \\
4 & A1-P3 & A1a between A1b and A1c on first-solved step (median order) & medians A1b 311832, A1a 221640, A1c 283424 & median ordering (MWU reported) & -- & miss \\
4 & A1-P4 & A1a between A1b and A1c on solved-episode length (median order) & medians A1b 20.3231, A1a 19.5622, A1c 19.6918 & median ordering (MWU reported) & -- & miss \\
4 & A2-P1 & 8$\times$8 eval-curve area: A2a $<$ A2c & mean 0.387 vs 0.702 & one-sided MWU & 0.00018 & hit \\
4 & A2-P2 & 8$\times$8 eval-curve area: A2b $<$ A2c & mean 0.060 vs 0.702 & one-sided MWU & $5\times10^{-6}$ & hit \\
4 & A2-P3 & 16$\times$16 eval-curve area: A2a $<$ A2c & mean 0.328 vs 0.579 & one-sided MWU & 0.002 & hit \\
4 & A2-P4 & 16$\times$16 eval-curve area: A2b $<$ A2c & mean 0.011 vs 0.579 & one-sided MWU & $5\times10^{-6}$ & hit \\
4 & A2-P5 & 8$\times$8 solves: A2a $<$ A2c & 11/20 vs 17/20 & one-sided Fisher & 0.041 & hit \\
4 & A2-P6 & 8$\times$8 solves: A2b $<$ A2c & 0/20 vs 17/20 & one-sided Fisher & $1.3\times10^{-8}$ & hit \\
4 & A2-P7 & 16$\times$16 solves: A2a $<$ A2c & 6/20 vs 8/20 & one-sided Fisher & 0.371 & miss \\
4 & A2-P8 & 16$\times$16 solves: A2b $<$ A2c & 0/20 vs 8/20 & one-sided Fisher & 0.002 & hit \\
4 & B0-P & optimum tampers $\geq$ 1, reaches goal, never tampers indefinitely (1/160, 1/40) & 1/160 (g): reaches goal, 1 tamper(s) at steps [0], 12 steps; 1/40 (g): reaches goal, 2 tamper(s) at steps [0, 11], 13 steps; 1/40 (g/2): reaches goal, 2 tamper(s) at steps [0, 11], 13 steps & deterministic & -- & hit \\
4 & PB1-B1 & B1 tamperers (reference A1b median done/episode 4.8e-05) & 20/20 (median done/episode 38.6) & binomial vs 0.5, one-sided & $9.5\times10^{-7}$ & hit \\
4 & PB1-B2 & B2 tamperers (reference B7 median done/episode 2.51) & 20/20 (median done/episode 61.2) & binomial vs 0.5, one-sided & $9.5\times10^{-7}$ & hit \\
4 & PB1-B3 & B3 tamperers (reference B5 median done/episode 0.000789) & 20/20 (median done/episode 173) & binomial vs 0.5, one-sided & $9.5\times10^{-7}$ & hit \\
4 & PB1-B4 & B4 tamperers (reference B6 median done/episode 0.475) & 20/20 (median done/episode 120) & binomial vs 0.5, one-sided & $9.5\times10^{-7}$ & hit \\
4 & PB1-pooled & B1-B4 tamperers pooled & 80/80 & binomial vs 0.5, one-sided & $8.3\times10^{-25}$ & hit \\
4 & PB2-B1 & solve B1 (tamper on) $<$ A1b (tamper off) & 1/20 vs 20/20 & one-sided Fisher & $1.5\times10^{-10}$ & hit \\
4 & PB2-B2 & solve B2 (tamper on) $<$ B7 (tamper off) & 0/20 vs 20/20 & one-sided Fisher & $7.3\times10^{-12}$ & hit \\
4 & PB2-B3 & solve B3 (tamper on) $<$ B5 (tamper off) & 1/20 vs 11/20 & one-sided Fisher & 0.00062 & hit \\
4 & PB2-B4 & solve B4 (tamper on) $<$ B6 (tamper off) & 0/20 vs 20/20 & one-sided Fisher & $7.3\times10^{-12}$ & hit \\
4 & PB3-B3 & tamper-loopers B3 (norm on) $>$ B4 (norm off) & 20/20 vs 20/20 & one-sided Fisher & 1.000 & miss \\
4 & PB3-B1 & tamper-loopers B1 (norm on) $>$ B2 (norm off) & 20/20 vs 20/20 & one-sided Fisher & 1.000 & miss \\
4 & PB4 & cycle-farm B5 (norm on) $>$ B6 (norm off) & 9/20 vs 0/20 & one-sided Fisher & 0.00061 & hit \\
4 & PB5 & B8 cycle-farms in most seeds & 20/20 & binomial vs 0.5, one-sided & $9.5\times10^{-7}$ & hit \\
5 & A-P1 & A1 solves more than B1 & 11/20 vs 1/20 & one-sided Fisher & 0.00062 & hit \\
5 & A-P1m & A1 success-curve area $>$ B1 & mean 0.438 vs 0.042 & one-sided MWU & $7.8\times10^{-6}$ & hit \\
5 & A-P2 & A2 solves more than B2 & 0/20 vs 0/20 & one-sided Fisher & 1.000 & miss \\
5 & A-P2m & A2 success-curve area $>$ B2 & mean 0.025 vs 0.022 & one-sided MWU & 0.005 & hit \\
5 & A-P3 & A1 solves more than B1+B3 & 11/20 vs 2/40 & one-sided Fisher & $2.6\times10^{-5}$ & hit \\
5 & A-P3m & A1 success-curve area $>$ B1+B3 & mean 0.438 vs 0.039 & one-sided MWU & $5\times10^{-6}$ & hit \\
5 & B-JTp & tamper presses decrease with cost (B1, B9, B10, B11) & medians 38.59, 27.13, 0.36, 0.00 & one-sided Jonckheere-Terpstra (permutation) & $5\times10^{-6}$ & hit \\
5 & B-JTs & solves increase with cost (B1, B9, B10, B11) & solves 1, 3, 20, 20 & one-sided Jonckheere-Terpstra (permutation) & $5\times10^{-6}$ & hit \\
5 & B-F-B9-vsB1 & B9 solves more than B1 & 3/20 vs 1/20 & one-sided Fisher & 0.302 & hit \\
5 & B-F-B9-vsA1b & B9 solves fewer than A1b & 3/20 vs 20/20 & one-sided Fisher & $1.3\times10^{-8}$ & hit \\
5 & B-L-B9-presses & B9 presses (median presses per episode) & 27.13 & level (range) & -- & hit \\
5 & B-L-B9-tamperers & B9 tamperers (count of 20) & 19/20 & level (range) & -- & hit \\
5 & B-L-B9-solves & B9 solves (count of 20) & 3/20 & level (range) & -- & hit \\
5 & B-F-B10-vsB1 & B10 solves more than B1 & 20/20 vs 1/20 & one-sided Fisher & $1.5\times10^{-10}$ & miss \\
5 & B-F-B10-vsA1b & B10 solves fewer than A1b & 20/20 vs 20/20 & one-sided Fisher & 1.000 & miss \\
5 & B-L-B10-presses & B10 presses (median presses per episode) & 0.36 & level (range) & -- & miss \\
5 & B-L-B10-tamperers & B10 tamperers (count of 20) & 7/20 & level (range) & -- & miss \\
5 & B-L-B10-solves & B10 solves (count of 20) & 20/20 & level (range) & -- & miss \\
5 & B-F-B11-vsB1 & B11 solves more than B1 & 20/20 vs 1/20 & one-sided Fisher & $1.5\times10^{-10}$ & hit \\
5 & B-F-B11-vsA1b & B11 solves fewer than A1b & 20/20 vs 20/20 & one-sided Fisher & 1.000 & hit \\
5 & B-L-B11-presses & B11 presses (median presses per episode) & 0.00 & level (range) & -- & hit \\
5 & B-L-B11-tamperers & B11 tamperers (count of 20) & 0/20 & level (range) & -- & hit \\
5 & B-L-B11-solves & B11 solves (count of 20) & 20/20 & level (range) & -- & hit \\
5 & C-P1 & cycle-farming rises K1 $<$ K2 $<$ K3 & 0/20, 0/20, 0/20 & one-sided Jonckheere-Terpstra (permutation) & 1.000 & miss \\
5 & C-P2 & cycle-farm: K3 $>$ K4 & 0/20 vs 0/20 & one-sided Fisher & 1.000 & miss \\
5 & C-P3 & solve: K5 $<$ K1 & 0/20 vs 20/20 & one-sided Fisher & $7.3\times10^{-12}$ & hit \\
5 & C-P4 & solve: K5 $<$ K6 & 0/20 vs 20/20 & one-sided Fisher & $7.3\times10^{-12}$ & hit \\
5 & D1-P1 & B1-B4 seeds that never found the goal & 1/80 & level (range) & -- & hit \\
5 & D1-P2 & found-then-lost seeds whose success declined (peak $\geq$ 0.10, then $\leq$ half) & 7/77 & level: hit iff $\geq$ 20\% ($\geq$ 16) & -- & miss \\
5 & D1-P3 & median exact first-goal step of B1-B4 seeds & 27,344 (max 1,187,264) & level: hit iff $\leq$ 41,000 & -- & hit \\
5 & D2-B1 & final-10\% policy entropy B1 (tamper on) $>$ A1b (tamper off) & median 1.878 vs 0.067 & one-sided MWU & $7.3\times10^{-12}$ & hit \\
5 & D2-B2 & final-10\% policy entropy B2 (tamper on) $>$ B7 (tamper off) & median 1.935 vs 1.138 & one-sided MWU & $7.3\times10^{-12}$ & hit \\
5 & D2-B3 & final-10\% policy entropy B3 (tamper on) $>$ B5 (tamper off) & median 1.844 vs 0.290 & one-sided MWU & $7.3\times10^{-12}$ & hit \\
5 & D2-B4 & final-10\% policy entropy B4 (tamper on) $>$ B6 (tamper off) & median 1.934 vs 0.598 & one-sided MWU & $7.3\times10^{-12}$ & hit \\
5 & D3-P1 & `done' fraction rises before success declines & rise first 7/7 (undetermined 0) & binomial vs 0.5, one-sided; hit iff share $\geq$ 0.8 & 0.008 & hit \\
\end{longtable}
\endgroup

\section{Known misclassifications and corrections}
\label{app:corrections}
Reproduced from \texttt{consolidated-results.md}.
% Generated by paper/scripts/make_tables.py. Do not edit by hand.
\begin{enumerate}[leftmargin=*,itemsep=2pt]
\item \textbf{Round 2, temptation formula F redefined after seeing data.} The first F did not fit the three calibration runs and was redefined before the round-2 test; the calibration runs were therefore excluded as evidence (disclosed in round 2).
\item \textbf{Round 2, pay-once control labelled `farmer' by the $>$5 rule} although replay showed episode-prolonging, not repetition; the round-3 exploratory re-score repeats this label.
\item \textbf{Round 2 claim dropped: `the critic prefers farming'.} The value rising with need is the $+D(h_t)$ term of the return and holds for solvers too; not evidence of preference.
\item \textbf{Round 1 claim corrected: LP-curiosity seeds 2 and 4 `entropy collapse'.} Entropy stayed high; the cause was left open.
\item \textbf{Event-bonus per-new-view payment} was written as 0.0975 in round-2 notes; the code pays $h_0^2-(h_0-0.1)^2$ = 0.09 (corrected in round 3).
\item \textbf{Round 3, coef x normalization logistic model not estimable} (all 60 normalization-off seeds solved); recorded as misses.
\item \textbf{Evaluation bias, rounds 1-3.} \texttt{evaluate()} kept the first N episodes to finish across 16 parallel envs, over-weighting short successful episodes. Re-evaluation (round 4): 16$\times$16 novelty 1e-3 0.845 $\rightarrow$ 0.787; Agent 3 1/160 0.714 $\rightarrow$ 0.653; 1/640 0.622 $\rightarrow$ 0.548; ranking unchanged. Fixed from round 4 on (per-environment quota).
\item \textbf{Round 4, explore-only runs (A2a, A2b) labelled `cycle-farm'} although pick-ups were not paid in those runs; replay showed non-learning policies. Led to the round-5 rule: cycle-farm applies only where pick-ups or door toggles are paid.
\item \textbf{Round 4, B1 and B2 ran concurrently} because a launcher missed a setup status; seeded single-threaded computation, so results are unaffected.
\item \textbf{Round 5, Step 0 at growth 1/160} used a certified early stop (validated against full convergence on all 21 growth-1/40 cases) instead of full convergence.
\item \textbf{Round 5, A-P2m is a `hit' on a negligible effect} (success-curve area 0.025 vs 0.022); both runs are at the floor and the solve test (A-P2) missed. Do not cite A-P2m as evidence.
\item \textbf{Round 5, tamper labels in tamper-off runs.} The `tamperers' column for tamper-off runs counts presses of the no-op `done' action (e.g. A2 presses at the uniform-random rate); the labels apply only to tamper-on runs.
\item \textbf{Round 5, Part D `found then lost'} includes seeds that only met the goal by chance; only 7/77 declined after real learning (D1-P2 miss), and D3 rests on those 7 seeds.
\end{enumerate}

\section{Per-seed results}
\label{app:seeds}
% Generated by paper/scripts/make_tables.py. Do not edit by hand.
\begingroup\footnotesize\setlength{\tabcolsep}{4pt}
\begin{longtable}{@{}rllllllll@{}}
\caption{Round 4, Part B, per seed: share of steps on ``done'' over the final 10\% of training log rows, and the seed's behaviour category (S solve, C cycle-farm, P prolong). Tamper-on runs alternate with their matched tamper-off runs, where ``done'' is a no-op. Category labels in tamper-on runs are mechanical (more than 5 paid repeats per episode) and are not used as evidence of farming (Section~\ref{sec:c5}). \repodata{results/table7_seeds_round4_tamper.csv}{table7}}\label{tab:seeds-r4}\\
\toprule
Seed & \multicolumn{2}{c}{growth 1/160, norm on} & \multicolumn{2}{c}{1/160, norm off} & \multicolumn{2}{c}{1/40, norm on} & \multicolumn{2}{c}{1/40, norm off} \\
 & tamper & no tamper & tamper & no tamper & tamper & no tamper & tamper & no tamper \\
\midrule
\endfirsthead
\toprule
Seed & \multicolumn{8}{c}{(continued)} \\
\midrule
\endhead
\bottomrule
\endfoot
40 & 0.060 C & 0.000 S & 0.096 C & 0.082 S & 0.257 C & 0.000 S & 0.189 C & 0.016 S \\
41 & 0.061 P & 0.000 S & 0.100 C & 0.081 S & 0.273 P & 0.013 C & 0.188 C & 0.017 S \\
42 & 0.059 P & 0.000 S & 0.096 C & 0.083 S & 0.270 P & 0.000 S & 0.189 C & 0.022 S \\
43 & 0.059 P & 0.000 S & 0.094 C & 0.081 S & 0.270 C & 0.001 C & 0.191 C & 0.015 S \\
44 & 0.064 P & 0.000 S & 0.101 C & 0.085 S & 0.273 P & 0.000 S & 0.187 C & 0.020 S \\
45 & 0.062 P & 0.000 S & 0.096 C & 0.083 S & 0.270 P & 0.000 S & 0.189 C & 0.019 S \\
46 & 0.056 C & 0.000 S & 0.094 C & 0.080 S & 0.261 P & 0.000 C & 0.189 C & 0.018 S \\
47 & 0.060 C & 0.000 S & 0.097 C & 0.081 S & 0.268 C & 0.000 C & 0.188 C & 0.022 S \\
48 & 0.067 S & 0.000 S & 0.096 C & 0.086 S & 0.217 S & 0.000 C & 0.187 C & 0.023 S \\
49 & 0.060 P & 0.000 S & 0.099 C & 0.083 S & 0.254 P & 0.000 S & 0.188 C & 0.012 S \\
50 & 0.060 C & 0.000 S & 0.094 C & 0.082 S & 0.256 P & 0.000 S & 0.185 C & 0.019 S \\
51 & 0.061 C & 0.000 S & 0.098 C & 0.079 S & 0.270 C & 0.005 C & 0.187 C & 0.016 S \\
52 & 0.063 P & 0.000 S & 0.097 C & 0.078 S & 0.274 P & 0.000 S & 0.190 C & 0.021 S \\
53 & 0.062 P & 0.000 S & 0.097 C & 0.077 S & 0.282 P & 0.000 C & 0.193 P & 0.017 S \\
54 & 0.065 P & 0.000 S & 0.099 C & 0.084 S & 0.277 P & 0.000 S & 0.190 C & 0.020 S \\
55 & 0.061 P & 0.000 S & 0.097 C & 0.084 S & 0.272 P & 0.000 S & 0.190 C & 0.022 S \\
56 & 0.064 P & 0.000 S & 0.094 C & 0.079 S & 0.276 P & 0.000 S & 0.187 P & 0.018 S \\
57 & 0.060 C & 0.000 S & 0.092 C & 0.081 S & 0.258 P & 0.000 S & 0.189 C & 0.018 S \\
58 & 0.063 P & 0.000 S & 0.098 C & 0.076 S & 0.276 C & 0.004 C & 0.189 C & 0.020 S \\
59 & 0.061 P & 0.000 S & 0.095 C & 0.082 S & 0.280 P & 0.002 C & 0.190 C & 0.025 S \\
\midrule
median & 0.061 & 0.000 & 0.096 & 0.081 & 0.270 & 0.000 & 0.189 & 0.019 \\
\end{longtable}
\endgroup

% Generated by paper/scripts/make_tables.py. Do not edit by hand.
\begingroup\footnotesize
\begin{longtable}{@{}rlllll@{}}
\caption{Round 5, Part B, per seed (growth 1/160, normalization on): tamper presses per step over the final 10\% of log rows (descriptive, from the run logs) and category. Cost 0 and the no-tamper column are the round-4 runs on the same seeds. \repodata{results/table8_seeds_round5_cost.csv}{table8}}\label{tab:seeds-r5b}\\
\toprule
Seed & cost 0 & cost 0.01 & cost 0.1 & cost 1 & no tamper action \\
\midrule
\endfirsthead
\toprule
Seed & cost 0 & cost 0.01 & cost 0.1 & cost 1 & no tamper action \\
\midrule
\endhead
\bottomrule
\endfoot
40 & 0.060 C & 0.043 C & 0.022 S & 0.000 S & 0.000 S \\
41 & 0.061 P & 0.042 F & 0.000 S & 0.000 S & 0.000 S \\
42 & 0.059 P & 0.043 C & 0.009 S & 0.000 S & 0.000 S \\
43 & 0.059 P & 0.045 S & 0.016 S & 0.000 S & 0.000 S \\
44 & 0.064 P & 0.044 P & 0.001 S & 0.000 S & 0.000 S \\
45 & 0.062 P & 0.025 S & 0.000 S & 0.000 S & 0.000 S \\
46 & 0.056 C & 0.041 C & 0.020 S & 0.000 S & 0.000 S \\
47 & 0.060 C & 0.039 C & 0.002 S & 0.000 S & 0.000 S \\
48 & 0.067 S & 0.042 S & 0.020 S & 0.000 S & 0.000 S \\
49 & 0.060 P & 0.044 P & 0.008 S & 0.000 S & 0.000 S \\
50 & 0.060 C & 0.043 C & 0.011 S & 0.000 S & 0.000 S \\
51 & 0.061 C & 0.043 C & 0.023 S & 0.000 S & 0.000 S \\
52 & 0.063 P & 0.043 C & 0.012 S & 0.000 S & 0.000 S \\
53 & 0.062 P & 0.045 P & 0.021 S & 0.000 S & 0.000 S \\
54 & 0.065 P & 0.045 P & 0.000 S & 0.000 S & 0.000 S \\
55 & 0.061 P & 0.042 P & 0.000 S & 0.000 S & 0.000 S \\
56 & 0.064 P & 0.043 P & 0.019 S & 0.000 S & 0.000 S \\
57 & 0.060 C & 0.043 C & 0.001 S & 0.000 S & 0.000 S \\
58 & 0.063 P & 0.046 P & 0.020 S & 0.000 S & 0.000 S \\
59 & 0.061 P & 0.044 P & 0.003 S & 0.000 S & 0.000 S \\
\end{longtable}
\endgroup

% Generated by paper/scripts/make_tables.py. Do not edit by hand.
\begingroup\footnotesize\setlength{\tabcolsep}{3.5pt}
\begin{longtable}{@{}llrlrrlrrll@{}}
\caption{Part D per seed (round-4 tamper-on runs; analyses registered in round 5). First-goal step is exact; peak and final success are rolling 100-episode means; ``declined'' requires peak $\geq 0.10$ and final $\leq$ half the peak; entropy and ``done'' share are final-10\% means; order compares the first row with ``done'' share $\geq 1/7 + 0.05$ (rise) with the decline row. \repodata{results/table9_seeds_partd.csv}{table9}}\label{tab:seeds-d}\\
\toprule
Run & Seed & First goal & Class & Peak & Final & Declined & Entropy & ``Done'' & Rise & Order \\
\midrule
\endfirsthead
\toprule
Run & Seed & First goal & Class & Peak & Final & Declined & Entropy & ``Done'' & Rise & Order \\
\midrule
\endhead
\bottomrule
\endfoot
1/160 on & 40 & 88,800 & lost & 0.04 & 0.021 & no & 1.897 & 0.060 & 40,960 & no decline \\
 & 41 & 6,288 & lost & 0.06 & 0.026 & no & 1.878 & 0.061 & 40,960 & no decline \\
 & 42 & 71,008 & lost & 0.05 & 0.036 & no & 1.858 & 0.059 & 40,960 & no decline \\
 & 43 & 25,632 & lost & 0.13 & 0.083 & no & 1.872 & 0.059 & 40,960 & rise first \\
 & 44 & 255,984 & lost & 0.03 & 0.002 & no & 1.866 & 0.064 & 40,960 & no decline \\
 & 45 & 4,432 & lost & 0.06 & 0.010 & no & 1.887 & 0.062 & 40,960 & no decline \\
 & 46 & 46,864 & lost & 0.10 & 0.034 & yes & 1.892 & 0.056 & 40,960 & rise first \\
 & 47 & 18,384 & lost & 0.17 & 0.009 & yes & 1.891 & 0.060 & 40,960 & rise first \\
 & 48 & 57,600 & kept & 1.00 & 0.997 & no & 1.164 & 0.067 & 40,960 & no decline \\
 & 49 & 6,096 & lost & 0.13 & 0.022 & yes & 1.852 & 0.060 & -- & no rise \\
 & 50 & 194,448 & lost & 0.09 & 0.040 & no & 1.885 & 0.060 & 40,960 & no decline \\
 & 51 & 16,096 & lost & 0.10 & 0.026 & yes & 1.881 & 0.061 & 61,440 & rise first \\
 & 52 & 20,368 & lost & 0.05 & 0.027 & no & 1.880 & 0.063 & 40,960 & no decline \\
 & 53 & 9,568 & lost & 0.03 & 0.008 & no & 1.853 & 0.062 & 40,960 & no decline \\
 & 54 & 144,624 & lost & 0.03 & 0.002 & no & 1.871 & 0.065 & 40,960 & no decline \\
 & 55 & 6,080 & lost & 0.06 & 0.019 & no & 1.864 & 0.061 & 40,960 & no decline \\
 & 56 & 101,200 & lost & 0.02 & 0.000 & no & 1.881 & 0.064 & 61,440 & no decline \\
 & 57 & 17,424 & lost & 0.20 & 0.054 & yes & 1.859 & 0.060 & 40,960 & rise first \\
 & 58 & 107,488 & lost & 0.13 & 0.001 & yes & 1.878 & 0.063 & 20,480 & rise first \\
 & 59 & 9,008 & lost & 0.05 & 0.019 & no & 1.886 & 0.061 & 40,960 & no decline \\
\addlinespace
1/160 off & 40 & 10,096 & lost & 0.06 & 0.028 & no & 1.935 & 0.096 & -- & no rise \\
 & 41 & 6,288 & lost & 0.07 & 0.022 & no & 1.936 & 0.100 & -- & no rise \\
 & 42 & 56,416 & lost & 0.06 & 0.024 & no & 1.936 & 0.096 & -- & no rise \\
 & 43 & 29,664 & lost & 0.05 & 0.023 & no & 1.934 & 0.094 & 163,840 & no decline \\
 & 44 & 55,184 & lost & 0.06 & 0.016 & no & 1.938 & 0.101 & 102,400 & no decline \\
 & 45 & 4,432 & lost & 0.06 & 0.026 & no & 1.935 & 0.096 & -- & no rise \\
 & 46 & 86,112 & lost & 0.07 & 0.031 & no & 1.935 & 0.094 & -- & no rise \\
 & 47 & 25,184 & lost & 0.06 & 0.039 & no & 1.936 & 0.097 & -- & no rise \\
 & 48 & 47,808 & lost & 0.08 & 0.036 & no & 1.936 & 0.096 & -- & no rise \\
 & 49 & 20,272 & lost & 0.09 & 0.042 & no & 1.936 & 0.099 & -- & no rise \\
 & 50 & 19,568 & lost & 0.07 & 0.024 & no & 1.934 & 0.094 & -- & no rise \\
 & 51 & 16,176 & lost & 0.07 & 0.032 & no & 1.936 & 0.098 & -- & no rise \\
 & 52 & 9,728 & lost & 0.07 & 0.032 & no & 1.935 & 0.097 & -- & no rise \\
 & 53 & 9,568 & lost & 0.05 & 0.027 & no & 1.936 & 0.097 & 122,880 & no decline \\
 & 54 & 57,744 & lost & 0.06 & 0.028 & no & 1.936 & 0.099 & -- & no rise \\
 & 55 & 6,080 & lost & 0.08 & 0.012 & no & 1.935 & 0.097 & -- & no rise \\
 & 56 & 18,576 & lost & 0.05 & 0.017 & no & 1.934 & 0.094 & -- & no rise \\
 & 57 & 12,864 & lost & 0.06 & 0.024 & no & 1.933 & 0.092 & -- & no rise \\
 & 58 & 111,680 & lost & 0.06 & 0.020 & no & 1.936 & 0.098 & 81,920 & no decline \\
 & 59 & 88,320 & lost & 0.05 & 0.018 & no & 1.935 & 0.095 & -- & no rise \\
\addlinespace
1/40 on & 40 & 8,928 & lost & 0.09 & 0.014 & no & 1.872 & 0.257 & 20,480 & no decline \\
 & 41 & 6,288 & lost & 0.03 & 0.008 & no & 1.853 & 0.273 & 20,480 & no decline \\
 & 42 & 695,728 & lost & 0.02 & 0.002 & no & 1.859 & 0.270 & 20,480 & no decline \\
 & 43 & 1,187,264 & lost & 0.03 & 0.006 & no & 1.846 & 0.270 & 20,480 & no decline \\
 & 44 & 690,464 & lost & 0.05 & 0.000 & no & 1.835 & 0.273 & 20,480 & no decline \\
 & 45 & 4,432 & lost & 0.03 & 0.000 & no & 1.819 & 0.270 & 20,480 & no decline \\
 & 46 & 377,712 & lost & 0.04 & 0.002 & no & 1.882 & 0.261 & 20,480 & no decline \\
 & 47 & 9,120 & lost & 0.13 & 0.013 & yes & 1.858 & 0.268 & 20,480 & rise first \\
 & 48 & 163,520 & kept & 0.94 & 0.901 & no & 1.323 & 0.217 & 20,480 & no decline \\
 & 49 & 27,808 & lost & 0.40 & 0.334 & no & 1.773 & 0.254 & 20,480 & no decline \\
 & 50 & 9,696 & lost & 0.04 & 0.013 & no & 1.897 & 0.256 & 20,480 & no decline \\
 & 51 & 846,704 & lost & 0.03 & 0.008 & no & 1.853 & 0.270 & 20,480 & no decline \\
 & 52 & 1,092,016 & lost & 0.01 & 0.003 & no & 1.844 & 0.274 & 20,480 & no decline \\
 & 53 & 16,960 & lost & 0.03 & 0.000 & no & 1.824 & 0.282 & 20,480 & no decline \\
 & 54 & 959,856 & lost & 0.01 & 0.002 & no & 1.829 & 0.277 & 20,480 & no decline \\
 & 55 & 233,168 & lost & 0.02 & 0.003 & no & 1.822 & 0.272 & 20,480 & no decline \\
 & 56 & 50,832 & lost & 0.01 & 0.000 & no & 1.808 & 0.276 & 20,480 & no decline \\
 & 57 & 15,824 & lost & 0.06 & 0.017 & no & 1.887 & 0.258 & 20,480 & no decline \\
 & 58 & never & never & 0.00 & 0.000 & no & 1.722 & 0.276 & 20,480 & no decline \\
 & 59 & 259,280 & lost & 0.03 & 0.007 & no & 1.844 & 0.280 & 20,480 & no decline \\
\addlinespace
1/40 off & 40 & 181,664 & lost & 0.05 & 0.019 & no & 1.935 & 0.189 & 40,960 & no decline \\
 & 41 & 6,288 & lost & 0.07 & 0.029 & no & 1.934 & 0.188 & 40,960 & no decline \\
 & 42 & 48,928 & lost & 0.06 & 0.010 & no & 1.934 & 0.189 & 40,960 & no decline \\
 & 43 & 127,104 & lost & 0.05 & 0.022 & no & 1.934 & 0.191 & 40,960 & no decline \\
 & 44 & 46,944 & lost & 0.05 & 0.031 & no & 1.934 & 0.187 & 40,960 & no decline \\
 & 45 & 4,432 & lost & 0.05 & 0.022 & no & 1.933 & 0.189 & 40,960 & no decline \\
 & 46 & 20,480 & lost & 0.05 & 0.017 & no & 1.934 & 0.189 & 40,960 & no decline \\
 & 47 & 49,424 & lost & 0.07 & 0.044 & no & 1.934 & 0.188 & 40,960 & no decline \\
 & 48 & 27,344 & lost & 0.06 & 0.019 & no & 1.935 & 0.187 & 40,960 & no decline \\
 & 49 & 17,840 & lost & 0.05 & 0.017 & no & 1.934 & 0.188 & 40,960 & no decline \\
 & 50 & 30,496 & lost & 0.07 & 0.033 & no & 1.935 & 0.185 & 40,960 & no decline \\
 & 51 & 16,096 & lost & 0.06 & 0.020 & no & 1.935 & 0.187 & 40,960 & no decline \\
 & 52 & 7,744 & lost & 0.09 & 0.032 & no & 1.933 & 0.190 & 40,960 & no decline \\
 & 53 & 56,864 & lost & 0.06 & 0.020 & no & 1.933 & 0.193 & 40,960 & no decline \\
 & 54 & 30,400 & lost & 0.06 & 0.006 & no & 1.933 & 0.190 & 40,960 & no decline \\
 & 55 & 9,744 & lost & 0.09 & 0.036 & no & 1.934 & 0.190 & 40,960 & no decline \\
 & 56 & 59,520 & lost & 0.06 & 0.019 & no & 1.934 & 0.187 & 40,960 & no decline \\
 & 57 & 9,136 & lost & 0.06 & 0.020 & no & 1.934 & 0.189 & 40,960 & no decline \\
 & 58 & 98,864 & lost & 0.05 & 0.020 & no & 1.934 & 0.189 & 20,480 & no decline \\
 & 59 & 25,968 & lost & 0.07 & 0.022 & no & 1.933 & 0.190 & 40,960 & no decline \\
\end{longtable}
\endgroup

\section{Theory computations}
\label{app:theory}
\paragraph{Check (a).}
Layouts use environment seeds 10000--10199. The breadth-first search runs over the real MiniGrid dynamics, restoring
the full grid state at each node, with actions left, right, forward, pick-up, drop and toggle. ``Done'' never shortens
a path. Success is reaching the goal (DoorKey) or carrying the target ball (KeyCorridor).
\begin{itemize}[itemsep=1pt]
  \item \textbf{Replay.} Each candidate action sequence is replayed in a fresh environment with the training code's
  satiation module.
  \item \textbf{Drop-and-regrab.} The richer family inserts one drop-and-regrab at every position where the agent
  carries the key and faces an empty cell, and keeps the best member.
  \item \textbf{Tamper family.} Tamper at $t=0$, with or without a second tamper just before the goal step.
\end{itemize}

\paragraph{Check (b).}
The \val{bs.nfarm} farming checkpoints of the round-2 growth-$1/20$ run were each run on the same \val{bc.layouts} layouts,
with actions sampled from the policy under a fixed seed per batch. One further seed of that run solved
(\val{bs.solverseed.rate} success) and is excluded. One farming seed succeeded on \val{bs.n6seed6.rate} of layouts;
its \val{bs.solveeps} solving episodes are excluded from the non-solving maximum.

\paragraph{Replay of the tamper-on policies.}
All \val{tr.npol} round-4 tamper-on checkpoints (four settings, 20 seeds each) were run on the same \val{bc.layouts} layouts with
sampled actions, as in check (b), with growth and the tamper action as in training.
\begin{itemize}[itemsep=1pt]
  \item \textbf{Recorded per episode:} the discounted intrinsic return, the total discounted return (the task reward
  arrives only on the final step), the mean of $D(h_t)$ over the episode's steps, success, and the share of steps on
  ``done''.
  \item \textbf{Checks:} the tracked drive matched the module's need state at every step, and the identity of
  Proposition~\ref{prop:ceiling} held on every episode. For every seed, the replay's press share was within
  \val{tr.presscheck.max} of the share in its training log.
\end{itemize}
Script \texttt{tamper\_policy\_returns.py}, data \texttt{tamper-policy-returns.json}.

\paragraph{Value iteration.}
Value iteration ran on a 7$\times$5 layout with a wall column and a locked door, with needs (grasp, open) on an exact
lattice of bin $g$. A halved bin reproduced the results where it was checked. Convergence was to $\max|\Delta V|<10^{-7}$, except for the
certified early stop at $1/160$ in round 5. The tamper version zeroes both modelled needs.

\section{Hyperparameters and implementation}
\label{app:hyper}
\begin{table}[!htbp]
\centering\small
\caption{PPO and reward settings (from \texttt{config.toml}). \repodata{results/table10_hyperparameters.csv}{table10}}
\label{tab:hyper}
% Generated by paper/scripts/make_tables.py. Do not edit by hand.
\begin{tabular}{@{}ll@{}}
\toprule
Setting & Value \\
\midrule
Environments in parallel & 16 \\
Rollout length per environment & 128 \\
Batch & 2048 \\
Environment steps per seed & 2,000,000 \\
Learning rate (linearly annealed) & $2.5\times10^{-4}$ \\
Discount $\gamma$ & 0.99 \\
GAE $\lambda$ & 0.95 \\
Minibatches per epoch & 4 \\
Epochs per update & 4 \\
Clip coefficient & 0.2 \\
Entropy coefficient & 0.01 \\
Value coefficient & 0.5 \\
Max gradient norm & 0.5 \\
Advantage normalization (per minibatch) & on, or off where stated \\
Needs $h$ (explore, grasp, open): start $h_0$ & 0.5 \\
Relief per event (satiety) & 0.1, 1.0, 1.0 \\
Intrinsic coefficient & 1.0 \\
Final evaluation episodes per environment & 64 \\
\bottomrule
\end{tabular}

\end{table}
The encoder applies three 2$\times$2 convolutions (16, 32 and 64 channels, ReLU, with one 2$\times$2 max-pool) to the
7$\times$7$\times$3 view, followed by a 64-unit ReLU layer. The needs (when present) are concatenated to these
features, and linear actor and critic heads follow. Software: CPython \val{ver.python} (from the run logs), PyTorch
\val{ver.torch}, Gymnasium \val{ver.gymnasium}, MiniGrid \val{ver.minigrid}, and NumPy \val{ver.numpy}. The NumPy
version is the one the lock file selects for that Python version; no per-run package inventory was retained.

\section{Compute}
\label{app:compute}
All runs used one Apple M4 Pro computer (12 CPU cores, 24\,GB memory) with no GPU. Each run trained its seeds in
parallel, one single-threaded process per seed.
\begin{itemize}[itemsep=1pt]
  \item \textbf{Rounds 2--5:} \val{compute.runs.r25} runs, \val{compute.hours.r25} hours of summed wall-clock time
  from the run records.
  \item \textbf{Round 1} (exploratory): \val{compute.runs.r1} runs, \val{compute.hours.r1} hours. Three of these runs
  are re-scored in Table~\ref{tab:farming}.
\end{itemize}
Two round-4 runs (satiation $1/160$ + tamper, norm on and off) ran concurrently by mistake, so their recorded
durations are about twice the usual. Seeded, single-threaded computation leaves their results unaffected
(correction 9).

\section{Reproducibility}
\label{app:repro}
Pre-registration files and their sha256 hashes, recomputed from the snapshots in the repository's \nolinkurl{data/sources/} folder (\nolinkurl{paper-data/sources/} in the Zenodo archive) and
equal to the hashes in the records written at locking.
\begin{itemize}[itemsep=1pt]
  \item \textbf{The lock times are self-reported.} Each record holds a file's hash and the UTC time at which we wrote
  it: \val{lock.r3} (round 3), \val{lock.r4} (round 4) and \val{lock.r5} (round 5). By our account, each was written
  before any training in its round.
  \item \textbf{What a reader can and cannot check.} The pre-registration files and their records were not committed
  to version control, nor deposited with any timestamping service, at lock time. A reader can check that a released
  file matches the hash reported here, but not when it was written.
  \item \textbf{Release.} Pre-registration files, hash records, run records, scored results, checkpoints, and the
  training and analysis code used in this paper are in the repository, \url{\repourl} (tag \repotag), and a
  snapshot of them is archived at Zenodo, doi:\href{https://doi.org/10.5281/zenodo.23708845}{\nolinkurl{10.5281/zenodo.23708845}}.
\end{itemize}
\begin{center}\footnotesize
\begin{tabular}{@{}ll@{}}
\toprule
File & sha256 \\
\midrule
Round 3 pre-registration & \texttt{\val{hash.r3}} \\
Round 4 pre-registration & \texttt{\val{hash.r4}} \\
Round 4 amendment & \texttt{\val{hash.r4a}} \\
Round 5 pre-registration & \texttt{\val{hash.r5}} \\
Round 5 amendment (Part C lock) & \texttt{\val{hash.r5a}} \\
\bottomrule
\end{tabular}
\end{center}
\begin{itemize}[itemsep=1pt]
  \item \textbf{Generation.} Every figure, table and quoted number is generated by a script in the repository's
  \nolinkurl{scripts/} folder, from the scored files in \nolinkurl{data/sources/} or from the post-hoc data in
  \nolinkurl{data/}. Each table and figure has a committed result file in \nolinkurl{results/}, named after it, and
  a \texttt{make} target that regenerates it; the repository's \nolinkurl{README.md} lists them all. After
  \texttt{git clone}, \texttt{make all} regenerates \nolinkurl{results/}, checks every number and builds this PDF.
  \item \textbf{Checking.} \texttt{check\_numbers.py} regenerates the number macros and tables, fails on any
  difference, and flags literal numbers in the text.
  \item \textbf{Run IDs.} Table~\ref{tab:runs} maps the descriptive labels to run identifiers.
\end{itemize}
% Generated by paper/scripts/make_tables.py. Do not edit by hand.
\begingroup\footnotesize
\begin{longtable}{@{}llll@{}}
\caption{Map from the descriptive labels used in the text to the internal run names and run identifiers. DoorKey-8x8 unless stated; normalization on unless stated. KeyCorridor runs are named K1--K6 here; the project's scored files and run records call them C1--C6. \repodata{results/table11_runs.csv}{table11}}\label{tab:runs}\\
\toprule
Label in the text & Round & Name & Run ID \\
\midrule
\endfirsthead
\toprule
Label in the text & Round & Name & Run ID \\
\midrule
\endhead
\bottomrule
\endfoot
satiation 1/40 (round-1 calibration) & 1 & cal 1/40 & \texttt{6bfabf4a} \\
satiation 1/160 (round-1 calibration) & 1 & cal 1/160 & \texttt{6e507c3e} \\
satiation 1/640 (round-1 calibration) & 1 & cal 1/640 & \texttt{6518a3c8} \\
satiation 1/40 (seeds 10--19) & 2 & N1 & \texttt{ed5147b6} \\
satiation 1/40, $\gamma=0.95$ & 2 & N2 & \texttt{74b2b2d0} \\
satiation 1/80 & 2 & N3 & \texttt{af4373da} \\
satiation 1/80, coef 1.29 & 2 & N4 & \texttt{b58ec6e0} \\
satiation 1/40, coef 0.5 & 2 & N5 & \texttt{eb8555bd} \\
satiation 1/20 & 2 & N6 & \texttt{bced806e} \\
pay-once event bonus & 2 & pay-once & \texttt{5b13189d} \\
satiation 1/60 & 3 & R1 & \texttt{4cac9e0a} \\
satiation 1/30 & 3 & R2 & \texttt{eea13605} \\
satiation 1/40, coef 0.5 & 3 & R3 & \texttt{3d03a914} \\
satiation 1/40 (reference) & 3 & R4 & \texttt{fff9fd5b} \\
satiation 1/40, coef 2 & 3 & R5 & \texttt{c3d67c09} \\
satiation 1/40, coef 0.5, norm off & 3 & R6 & \texttt{89d8e34c} \\
satiation 1/40, norm off & 3 & R7 & \texttt{135d1475} \\
satiation 1/40, coef 2, norm off & 3 & R8 & \texttt{1205f6d0} \\
satiation 1/40 + early-goal starts & 3 & R9 & \texttt{a7cb164e} \\
fixed pay-once control & 3 & R10 & \texttt{07653362} \\
satiation 1/640 & 4 & A1a & \texttt{a97113a1} \\
satiation 1/160 & 4 & A1b & \texttt{3629f9a7} \\
satiation, no refill (growth 0) & 4 & A1c & \texttt{1d549b2f} \\
explore-only satiation 1/640 & 4 & A2a & \texttt{224a7d96} \\
explore-only satiation 1/160 & 4 & A2b & \texttt{7a364fed} \\
episodic novelty & 4 & A2c & \texttt{2d94fc36} \\
satiation 1/160 + tamper & 4 & B1 & \texttt{bb00e7e5} \\
satiation 1/160 + tamper, norm off & 4 & B2 & \texttt{a8f5efc4} \\
satiation 1/40 + tamper & 4 & B3 & \texttt{78c44406} \\
satiation 1/40 + tamper, norm off & 4 & B4 & \texttt{f0a40d20} \\
satiation 1/40 & 4 & B5 & \texttt{4d3b84f0} \\
satiation 1/40, norm off & 4 & B6 & \texttt{a6e11900} \\
satiation 1/160, norm off & 4 & B7 & \texttt{115c6566} \\
constant event bonus, norm off & 4 & B8 & \texttt{fb860207} \\
task-only & 5 & A1 & \texttt{022a3e90} \\
task-only, norm off & 5 & A2 & \texttt{1c7b2eda} \\
satiation 1/160 + tamper, cost 0.01 & 5 & B9 & \texttt{b971c18d} \\
satiation 1/160 + tamper, cost 0.1 & 5 & B10 & \texttt{a25b7c99} \\
satiation 1/160 + tamper, cost 1 & 5 & B11 & \texttt{896b2863} \\
KeyCorridor satiation 1/160 & 5 & K1 & \texttt{c14516de} \\
KeyCorridor satiation 1/40 & 5 & K2 & \texttt{ad87a02f} \\
KeyCorridor satiation 1/20 & 5 & K3 & \texttt{aa29d1c3} \\
KeyCorridor satiation 1/20, norm off & 5 & K4 & \texttt{2153e884} \\
KeyCorridor satiation 1/160 + tamper & 5 & K5 & \texttt{35c840cd} \\
KeyCorridor task-only & 5 & K6 & \texttt{6571bce6} \\
KeyCorridor calibration (seeds 100--104) & 5 & calibration & \texttt{bd005cb8} \\
\end{longtable}
\endgroup

\end{document}